\documentclass[10pt,journal,compsoc]{IEEEtran}
\ifCLASSOPTIONcompsoc
  \usepackage[nocompress]{cite}
\else
  \usepackage{cite}
\fi

\usepackage[pdftex]{graphicx}
\graphicspath{{Figures/}}
\usepackage{amsmath}
\usepackage{amssymb}
\ifCLASSOPTIONcompsoc
  \usepackage[caption=false,font=footnotesize,labelfont=sf,textfont=sf]{subfig}
\else
  \usepackage[caption=false,font=footnotesize]{subfig}
\fi
\usepackage[font=small]{caption}

\usepackage{bm}
\newcommand{\matr}[1]{\mathbf{#1}}

\usepackage[flushleft]{threeparttable}
\usepackage{url}
\usepackage{xcolor}

\begin{document}
\title{Deep Independently Recurrent Neural Network (IndRNN)}

\author{Shuai~Li,
        Wanqing~Li,~\IEEEmembership{Senior Member,~IEEE,}
        Chris~Cook,
        Yanbo~Gao

\IEEEcompsocitemizethanks{
\IEEEcompsocthanksitem S. Li and Y. Gao are with Shandong University, Jinan, China. Email: \{shuaili, ybgao\}@sdu.edu.cn.\IEEEcompsocthanksitem W. Li and C. Cook are with University of Wollongong, NSW 2522, Australia. E-mail: \{wanqing, ccook\}@uow.edu.au
}
\thanks{Manuscript received October 10, 2019.}}

\IEEEtitleabstractindextext{%
\begin{abstract}
Recurrent neural networks (RNNs) are known to be difficult to train due to the gradient vanishing and exploding problems and thus difficult to learn long-term patterns and construct deep networks. To address these problems, this paper proposes a new type of RNNs with the recurrent connection formulated as Hadamard product, referred to as independently recurrent neural network (IndRNN), where neurons in the same layer are independent of each other and connected across layers. Due to the better behaved gradient backpropagation, IndRNN with regulated recurrent weights effectively addresses the gradient vanishing and exploding problems and thus long-term dependencies can be learned. Moreover, an IndRNN can work with non-saturated activation functions such as ReLU (rectified linear unit) and be still trained robustly. Different deeper IndRNN architectures, including the basic stacked IndRNN, residual IndRNN and densely connected IndRNN, have been investigated, all of which can be much deeper than the existing RNNs. Furthermore, IndRNN reduces the computation at each time step and can be over 10 times faster than the commonly used Long short-term memory (LSTM). Experimental results have shown that the proposed IndRNN is able to process very long sequences and construct very deep networks. Better performance has been achieved on various tasks with IndRNNs compared with the traditional RNN, LSTM and the popular Transformer. 
\end{abstract}

}

\maketitle

\IEEEdisplaynontitleabstractindextext

\IEEEpeerreviewmaketitle

\IEEEraisesectionheading{\section{Introduction}\label{sec:introduction}}

\IEEEPARstart{L}{ong-term} dependency is important for many applications. Especially for applications processing temporal sequences such as action recognition \cite{Liu2018SkeletonBasedAR, Zhang2018ViewAN} and language processing \cite{Zhang2016DrawingAR,Wu2016ImageCA}, the past information is important for the recognition of the future events. There are also applications exploring spatial context information such as scene segmentation \cite{Shuai2018SceneSW} and spatial pooling \cite{li2017fully}. To explore the long-term dependency, recurrent neural networks (RNNs) \cite{jordan1997serial} have been widely used and have achieved impressive results. Compared with the feed-forward networks such as the convolutional neural networks (CNNs), a recurrent connection is added where the hidden state at the previous time step is used as an input to obtain the current state, in order to keep memory of the past information. The update of the hidden states at each time step follows:
\begin{equation}
\mathbf{h}_t=\sigma(\mathbf{Wx}_t+\mathbf{Uh}_{t-1}+\mathbf{b})
\label{RNN}
\end{equation}
where $\mathbf{x}_t\in \mathbb{R}^{M}$ and $\mathbf{h}_t\in \mathbb{R}^{N}$ are the input and hidden state at time step $t$, respectively. $\mathbf{W}\in \mathbb{R}^{N\times M}$, $\mathbf{U}\in \mathbb{R}^{N\times N}$ and $\mathbf{b}\in \mathbb{R}^{N}$ are the weights for the current input and the recurrent input, and the bias of the neurons, respectively. $\sigma$ is an element-wise activation function of the neurons, and $M$ and $N$ are the dimension of the input and the number of neurons in this RNN layer, respectively.

Due to the recurrent connections with repeated multiplication of the recurrent weight matrix, training of the RNNs suffers from the gradient vanishing and exploding problem. Despite the efforts in initialization and training techniques \cite{neyshabur2016path, krueger2016regularizing, le2015simple, talathi2015improving}, it is still very difficult to learn long-term dependency. Several RNN variants such as the long short-term memory (LSTM) \cite{greff2017lstm, Hochreiter1991UntersuchungenZD, jozefowicz2015empirical} and the gated recurrent unit (GRU) \cite{cho2014learning} have been proposed to address the gradient problems. However, the use of the hyperbolic tangent and the sigmoid functions as the activation function in these variants results in gradient decay over layers. While some researches investigate using deep RNNs with LSTM\cite{Kim2017ResidualLD, Li2018LayerTL}, much research has only focused on using some relative shallow RNNs such as in \cite{krueger2016zoneout, merity2018regularizing, shahroudy2016ntu, Liu2018SkeletonBasedAR, song2017end, zhang2017geometric}.  

On the other hand, the existing RNN models share the same component $\sigma(\mathbf{Wx}_t+\mathbf{Uh}_{t-1}+\mathbf{b})$ in (\ref{RNN}), where the recurrent connection connects all the neurons through time. This makes it hard to interpret and understand the roles of each individual neuron (e.g., what patterns each neuron responds to) without considering the others. Moreover, with the recurrent connections, matrix product is performed at each time step and the computation cannot be easily paralleled, leading to a very time-consuming process when dealing with long sequences.

In this paper, we propose a new type of RNN, referred to as independently recurrent neural network (IndRNN). In the proposed IndRNN, the recurrent inputs are processed with the Hadamard product as $\mathbf{h}_t=\sigma(\mathbf{Wx}_t+\mathbf{u}\odot\mathbf{h}_{t-1}+\mathbf{b})$. This provides a number of advantages over the traditional RNNs including:

\begin{itemize}
\item Able to process longer sequences: the gradient vanishing and exploding problem is effectively addressed by regulating the recurrent weights, and long-term memory can be kept in order to process long sequences. Experiments have demonstrated that an IndRNN can well process sequences over $5000$ steps.
\item Able to construct deeper networks: multiple layers of IndRNNs can be efficiently stacked, especially with skip-connection and dense connection, to increase the depth of the network. An example of 21-layer residual IndRNN and deep densely connected IndRNN are demonstrated in the experiments.
\item Able to be robustly trained with non-saturated functions such as ReLU: with the gradient backpropagation through time better behaved, non-saturated function such as ReLU\cite{Nair2010RectifiedLU} can be used as the activation function and be trained robustly. IndRNN with ReLU is used throughout the experiments.
\item Able to interpret the behaviour of IndRNN neurons independently without the effect from the others: since the neurons in one layer are independent of each other, each neuron's behaviour can be interpreted individually. Moreover, the relationship between the range of the memories and the recurrent weights are established through gradient backpropagation, and the memories learned by the task can be understood by visualizing the recurrent weights, as illustrated in experiments. 
\item Reduced complexity. With the new recurrent connections based on element-wise vector product, which is much more efficient than the matrix product, the complexity of IndRNN is greatly reduced compared with the traditional RNNs (over 10 times faster than the cuDNN LSTM). 
\end{itemize}

Experiments have demonstrated that IndRNN performs much better than the traditional RNN, LSTM  and Transformer models on the tasks of the adding problem, sequential MNIST classification, language modelling and action recognition. With the advantages brought by IndRNN, we are able to further show:
\begin{itemize}
\item Better performance can be achieved with deeper IndRNN architectures as verified for the sequential MNIST classification, language modelling and skeleton-based action recognition tasks.
\item Better performance can be achieved by learning with longer dependency as verified for the language modelling tasks.
\end{itemize}

Part of this paper has appeared in the conference paper \cite{li2018independently} where IndRNN is introduced and verified on some tasks without further analysing its advantage. Significant extension has been made in this paper. 1) New deep IndRNN architecture, densely connected IndRNN is proposed to enhance the feature reuse in addition to the residual IndRNN architecture.  2) The relationship between memory and recurrent weight is established through gradient backpropagation, and the learned memories are visualized for skeleton-based action recognition as an example. 3) More experiments are added including a new task, i.e., word-level language modelling. 4) Experiments are conducted to verify that IndRNN with longer temporal dependency and deeper architecture can achieve better performance. 5) Comparison has been made to the popular Transformer \cite{Vaswani2017AttentionIA,Wang2019RTransformerRN} on sequential MNIST classification, Char- and Word-level language modelling and skeleton based action recognition. 6) A faster implementation with CUDA optimization is provided and made publicly available, which can be over 10 times faster than the cuDNN LSTM.

The rest of this paper is organized as follows. Section \ref{sec_relatedwork} describes the related work in the literature. Section \ref{sec_indrnn} presents the proposed IndRNN with its gradient backpropagation through time process. It also describes the relationship between the recurrent weight and memory, and its complexity compared with the existing methods. Section \ref{sec_deepindrnn} explains different deep IndRNN architectures and Section \ref{sec_experiment} presents the experimental results. Finally, conclusions are drawn at Section \ref{sec_conclusion}

\section{Related Work}
\label{sec_relatedwork}
It is known that a simple RNN suffers from the gradient vanishing and exploding problem due to the repeated multiplication of the recurrent weight matrix, which makes it very difficult to train and capture long dependencies. In order to solve the gradient vanishing and exploding problem, long short-term memory (LSTM) \cite{hochreiter1997long} was introduced, with a constant error carousel (CEC) to enforce a constant error flow through time. Multiplicative gates including input gate, output gate and forget gate are employed to control the information flow, resulting in many more parameters than the simple RNN. A well known LSTM variant is the gated recurrent unit (GRU) \cite{cho2014learning} composed of a reset gate and an update gate, which reduces the number of parameters to some extent. It has been reported in various papers \cite{jozefowicz2015empirical} that GRU achieves similar performance as LSTM. There are also some other LSTM variants \cite{jozefowicz2015empirical, zhou2016minimal, collins2016capacity, greff2017lstm} reported in the literature. However, these architectures \cite{jozefowicz2015empirical, zhou2016minimal, collins2016capacity, greff2017lstm} generally take a similar form as LSTM and show a similar performance as well, so they are not discussed further. LSTM and its variants use gates on the input and the recurrent input to regulate the information flow through the network. However, the use of gates based on the recurrent input prevents parallel computation and thus increases the computational complexity of the whole network. To reduce the complexity and process the states of the network over time in parallel, QRNN (Quasi-Recurrent Neural Network) \cite{bradbury2016quasi} and SRU (Simple Recurrent Unit) \cite{lei2017training} were proposed where the recurrent connections are fixed to be identity connection and controlled by gates with only input information, thus making most of the computation parallel. While this strategy greatly simplifies the computational complexity, it reduces the capability of their RNNs since the recurrent connections are no longer trainable. By contrast, the proposed IndRNN with regulated recurrent weights addresses the gradient exploding and vanishing problems without losing the power of trainable recurrent connections and without involving gate parameters. Moreover, IndRNN reduces computation and runs much faster than LSTM and even SRU \cite{lei2017training}.

There are also some simple RNN variants trying to solve the gradient vanishing and exploding problem by altering the recurrent connections. In \cite{campos2017skip}, a skip RNN was proposed where a binary state update gate is added to control the network to skip the processing of one time step. In this way, fewer time steps may be processed and the gradient vanishing and exploding problem can be alleviated to some extent. In \cite{arjovsky2015unitary, wisdom2016full}, a unitary evolution RNN was proposed where the recurrent weights are empirically defined in the form of a unitary matrix. In this case, the norm of the backpropagated gradient can be bounded without exploding. In \cite{zhang2018learning}, a Fourier Recurrent Unit (FRU) was proposed where the hidden states over time are summarized with Fourier basis functions and the gradients are bounded. However, such methods usually introduce other transforms on the recurrent weight which complicates the recurrent units, making them hard to use and interpret. On the contrary, the proposed IndRNN further simplifies the recurrent connections and makes all the neurons in one layer independent from others, thus easier to interpret (e.g. understanding the long and short memories kept by the task).

There are also attempts at using the non-saturated activation functions such as ReLU (rectified linear unit) to reduce the gradient decay problem introduced by the activation function. While this reduces the gradient decay problem, it greatly aggravates the effect introduced by the repeated use of the recurrent weight matrix, making the network highly unstable. To alleviate this problem, works on initialization and training techniques, such as initializing the recurrent weights to a proper range or regulating the norm of the gradients over time, were proposed in the literature. In \cite{le2015simple}, an initialization technique was proposed for an RNN with ReLU activation, termed as IRNN, which initializes the recurrent weight matrix to be the identity matrix and bias to be zero. In \cite{talathi2015improving}, the recurrent weight matrix was further suggested to be a positive definite matrix with the highest eigenvalue being unity and all the remaining eigenvalues less than 1. In \cite{neyshabur2016path}, the geometry of RNNs was investigated and a path-normalized optimization method for training was proposed for RNNs with ReLU activation. In \cite{krueger2016regularizing}, a penalty term on the squared distance between successive hidden states' norms was proposed to prevent the exponential growth of IRNN's activation. Although these methods help ease gradient exploding, they are not able to completely avoid the problem (e.g. the eigenvalues of the recurrent weight matrix may still be larger than 1 in the process of training and lead to gradient exploding). Moreover, the training of an RNN with ReLU is very sensitive to the learning rate. When the learning rate is large, the gradient is likely to explode. The proposed IndRNN addresses the gradient problems by regulating the recurrent weights, which are then mapped to the constraint on the maximum gradient. It can work with ReLU and be trained robustly. As a result, an IndRNN is able to process very long sequences (e.g. over $5000$ steps as demonstrated in the experiments).

In addition to solving the gradient vanishing and exploding problem in order to process long sequences and learn long-term dependency, efforts are also put into investigating deeper RNN networks. Compared with the deep CNN architectures which could be over 100 layers such as the residual CNN \cite{he2016deep} and the densely connected CNN \cite{huang2017densely}, most of the existing RNN architectures only consist of several layers (2 or 3 for example \cite{krueger2016zoneout,shahroudy2016ntu,le2015simple}). This is partly due to the gradient vanishing and exploding problem which already results in the difficulty in training a single-layer RNN. A deeper RNN \cite{pascanu2013construct, chung2015gated} may further aggravate this problem. For LSTM models, since all the gate functions, input and output modulations employ sigmoid or hyperbolic tangent functions as the activation function, it suffers from the gradient decay over layers when multiple LSTM layers are stacked into a deep model. Currently, a few models were reported that employ residual connections \cite{he2016deep} between LSTM layers to make the network deeper \cite{wu2016google}. As shown in \cite{veit2016residual}, residual networks behave like ensembles of relatively shallow networks. For the residual connections to work efficiently, each residual block desires a good gradient behaviour. Due to gradient decay problem in each LSTM, the deep LSTM model with the residual connections cannot significantly improve the performance, which is also observed in \cite{pradhanexploring}. In \cite{zilly2016recurrent}, a recurrent highway network (RHN) was proposed where at each time step, multiple layers with highway connections are used to process the recurrent state. Compared with the LSTM, the transition process is greatly deepened. Recently, Transformers \cite{Vaswani2017AttentionIA} have been proposed using self-attention. It does not contain any recurrent connections, hence resembles a CNN more than a RNN. All in all, RNN architectures that can be stacked with multiple layers and efficiently trained are still highly desired. The proposed IndRNN well addresses this issue and can construct very deep models with the use of ReLU, residual or dense connections (e.g. over $20$ layers as demonstrated in the experiments).

Other than the above works making RNN process long sequences and construct deep networks, there are also other variants such as the multiplicative integration \cite{sutskever2011generating,wu2016multiplicative}, multidimensional \cite{kalchbrenner2015grid, Graves2007MRN} or bidirectional extensions, which is beyond the scope of this paper and thus not further explained. One interesting result reported in \cite{collins2016capacity} shows that the capacities of existing RNNs are very similar and the difference in their performances are driven primarily by differences in training effectiveness. Compared with the existing RNNs, the proposed IndRNN uses a simple formulation, and is easy and robust to train with different kinds of techniques such as the ReLU, batch normalization and dropout from CNN.

\begin{figure*}[tbp]
	\centering
	\subfloat[The conventional simple RNN with recurrent connections connecting all neurons over time.]{
	\includegraphics[width=0.35\hsize]{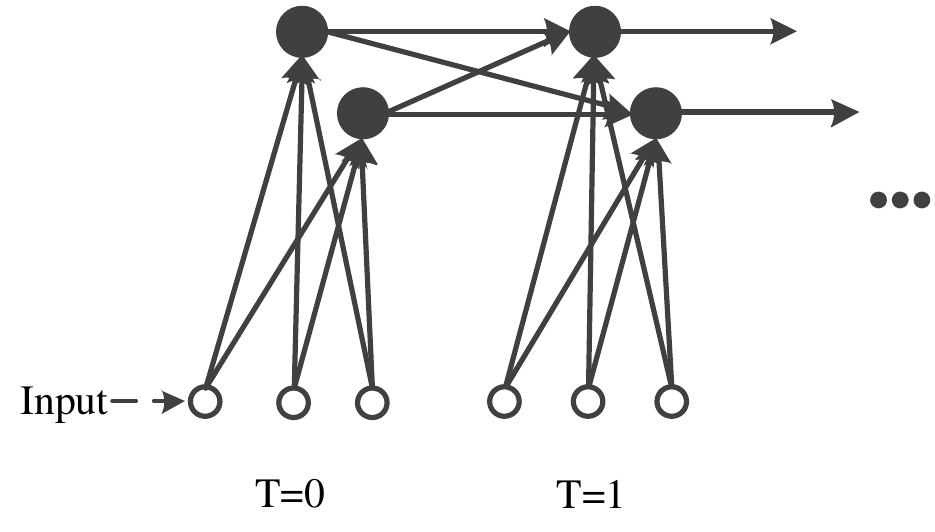}\label{img_rnn}} \hspace{2cm}
	\subfloat[The proposed IndRNN with independent recurrent connections making each neuron in one layer independent while connected over layers.]{
	\includegraphics[width=0.35\hsize]{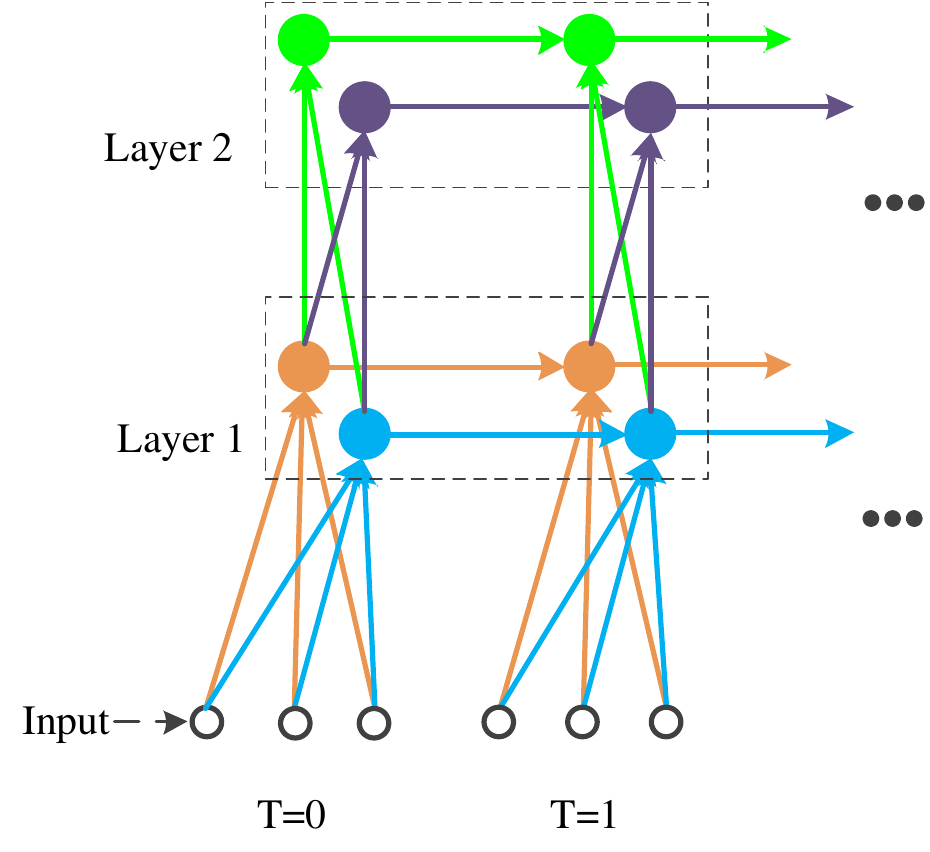}\label{img_indrnn}}
	\caption{Illustration of a conventional simple RNN and the proposed IndRNN unfolded in time. Each solid dot represents a neuron in a layer and each line represents a connection.} \label{fig_rnn_indrnn}
\end{figure*}

\section{Independently Recurrent Neural Network (IndRNN)}
\label{sec_indrnn}
The proposed independently recurrent neural network (IndRNN) follows:
\begin{equation}
\mathbf{h}_t=\sigma(\mathbf{Wx}_t+\mathbf{u}\odot\mathbf{h}_{t-1}+\mathbf{b})
\label{IndRNN}
\end{equation}
where $\mathbf{x}_t\in \mathbb{R}^{M}$ and $\mathbf{h}_t\in \mathbb{R}^{N}$ are the input and hidden state at time step $t$, respectively. $\mathbf{W}\in \mathbb{R}^{N\times M}$, $\mathbf{u}\in \mathbb{R}^{N}$ and $\mathbf{b}\in \mathbb{R}^{N}$ are the weights for the current input and the recurrent input, and the bias of the neurons, respectively. $\odot$ is the Hadamard product, $\sigma$ is an element-wise activation function of the neurons, and $N$ is the number of neurons in this RNN layer. Compared with the traditional RNN where the recurrent weight $\mathbf{U}$ is a matrix and processes the recurrent input using matrix product, the recurrent weight $\mathbf{u}$ in IndRNN is a vector and processes the recurrent input with element-wise vector product. Each neuron in one layer is independent from others, thus termed as ``independently recurrent''. For the $n$-th neuron, the hidden state $h_{n,t}$ can be obtained as 
\begin{equation}
h_{n,t}=\sigma(\mathbf{w}_{n}\mathbf{x}_t+u_nh_{n,t-1}+b_n)
\label{kernel_form}
\end{equation}
where $\mathbf{w}_{n}$, $u_n$ and $b_n$ are the $n$-th row of the input weight, recurrent weight and bias, respectively. The proposed IndRNN can be extended to a convolutional IndRNN where, instead of processing input of each time step using a fully connected weight ($\mathbf{Wx}_t$), it is processed with convolutional operation ($\mathbf{W*x}_t$, where $*$ denotes the convolution operator).

Fig. \ref{img_indrnn} illustrates the proposed IndRNN unfolded in time. Each solid dot represents a neuron in a layer and each line represents a connection. It can be seen that the processing of each neuron in one layer is independent from each other as noted by different colors. Each neuron only receives information from the input and its own hidden state at the previous time step. That is, each neuron in an IndRNN deals with one type of spatial-temporal pattern independently. On the contrary, neurons in the simple RNN are connected together over time by the recurrent weight connection as shown in Fig. \ref{img_rnn}. Fig. \ref{img_indrnn} also shows that one more IndRNN layer can explore the correlation (cross-channel information) among the neurons in one layer as the neuron in the following layer takes output of all the neurons in the previous layer as input. The relationship between IndRNN and the conventional RNN is illustrated in the Appendix \ref{appenx_a}, where we have shown that under certain circumstances (e.g. linear activation), a traditional RNN is a special case of a two-layer IndRNN.

\begin{table*}
\centering
\caption{Relationship between recurrent weight and memory through gradient.} 
\label{ch5_weight_memory}
\setlength{\tabcolsep}{2em}
  \begin{tabular}{c  c  c}
  \hline
   \textbf{Memory} & \textbf{Gradient} & \textbf{Recurrent weight} \\
  \hline
  Long-term & effective after $T-t$ steps: $\frac{\partial {J_{n,T}}}{\partial {h}_{n,t}}>\epsilon$ & $|u_n|\in (\sqrt[\leftroot{-3}\uproot{5}(T-t)]{\frac{\epsilon}{|\prod_{k=t}^{T-1} {\sigma'}_{n,k+1}|}}, +\infty)$ \\
  \hline
  Short-term & reduced after $T-t$ steps: $\frac{\partial {J_{n,T}}}{\partial {h}_{n,t}}\leq \epsilon$ & $|u_n|\in [0, \sqrt[\leftroot{-3}\uproot{5}(T-t)]{\frac{\epsilon}{|\prod_{k=t}^{T-1} {\sigma'}_{n,k+1}|}}]$ \\
  \hline
  & not exploding: $\frac{\partial {J_{n,T}}}{\partial {h}_{n,t}}\leq\gamma$ & $|u_n|\in [0, \sqrt[\leftroot{-3}\uproot{5}(T-t)]{\frac{\gamma}{|\prod_{k=t}^{T-1} {\sigma'}_{n,k+1}|}}]$\\
  \hline
  \end{tabular}
\end{table*}

\subsection{Backpropagation Through Time for An IndRNN}
\label{BPTT}
Since the neurons in each layer are independent of each other, the gradient backpropagation through time for neurons in one layer can also be performed independently. For the $n$-th neuron $h_{n,t}=\sigma(\mathbf{w}_{n}\mathbf{x}_t+u_nh_{n,t-1})$ where the bias is ignored, suppose the objective trying to minimize at time step $T$ is $J_n$. Then the gradient back propagated to the time step $t$ is 
\begin{small}
{
\begin{align}
\frac{\partial {J_n}}{\partial {h}_{n,t}} &= \
\frac{\partial {J_n}}{\partial {h}_{n,T}}\frac{\partial {h}_{n,T}}{\partial {h}_{n,t}}  
=\frac{\partial {J_n}}{\partial {h}_{n,T}}\
\prod_{k=t}^{T-1} \frac{\partial {h}_{n,k+1}}{\partial {h}_{n,k}} \nonumber \\
&=\frac{\partial {J_n}}{\partial {h}_{n,T}} \prod_{k=t}^{T-1} {\sigma'}_{n,k+1} {u}_n \ 
=\frac{\partial {J_n}}{\partial {h}_{n,T}} {u}_n^{T-t} \prod_{k=t}^{T-1} {\sigma'}_{n,k+1} \
\label{KIRNNgradient}
\end{align}
}
\end{small}
where ${\sigma'}_{n,k+1}$ is the derivative of the element-wise activation function. It can be seen that the gradient only involves the exponential term of a scalar value $u_n$ which can be easily regulated, and the gradient of the activation function which is often bounded in a certain range. Compared with the gradients of an RNN ($\frac{\partial J}{\partial h_T} \prod_{k=t}^{T-1} diag(\sigma'(h_{k+1})) \matr{U}^T $ where $diag(\sigma'(h_{k+1}))$ is the Jacobian matrix of the element-wise activation function), the gradient of an IndRNN directly depends on the value of the recurrent weight (which is changed by a small magnitude according to the learning rate) instead of matrix product (which is mainly determined by its eigenvalues and can be changed significantly even though the change to each matrix entries is small \cite{parlett1964laguerre}). Thus the training of an IndRNN is more robust than a traditional RNN. 

To address the gradient exploding and vanishing problem over time, we need to regulate the exponential term ``${u}_n^{T-t} \prod_{k=t}^{T-1} {\sigma'}_{n,k+1}$'' to an appropriate range (ignoring the gradient backpropagated from the objective to the hidden state at time step $T$, $\frac{\partial {J_n}}{\partial {h}_{n,T}}$). Note that for some activation functions such as ReLU, when not activated, its derivative is $0$ and the neuron cannot learn anything. Here we only discuss the gradient when neurons are activated ($\prod_{k=t}^{T-1} {\sigma'}_{n,k+1}\neq 0$). For gradient exploding, denoting the  magnitude of the largest gradient value without exploding by $\gamma$, the constraint ``$|{u}_n^{T-t} \prod_{k=t}^{T-1} {\sigma'}_{n,k+1}|\leq\gamma$'' avoids the gradient exploding problem. Accordingly, the recurrent weight $|u_n|$ satisfies $|u_n|\leq \sqrt[\leftroot{-3}\uproot{5}(T-t)]{\frac{\gamma}{|\prod_{k=t}^{T-1} {\sigma'}_{n,k+1}|}}$. For the commonly used activation functions such as ReLU and tanh, their derivatives are no larger than $1$, i.e., $|{\sigma'}_{n,k+1}|\leq 1$. Especially for ReLU, its gradient $\sigma'$ is either $0$ or $1$. Therefore, this constraint can be simplified to $|u_n|\leq \sqrt[\leftroot{-3}\uproot{5}(T-t)]{\gamma}$. For especially long sequences, it can be further simplified to $|u_n|\leq 1$. 

On the other hand, for the gradient vanishing problem, the magnitude of the gradient needs to be larger than $0$, which can be met by $|u_n| > 0$. In the literature and practice, when gradient vanishing is referred, usually we expect the gradient after many time steps (i.e. $T-t$) to be larger than a small value ($\epsilon$) in order for the gradient to effectively make a change. Therefore, we further strength the above gradient constraint to $|{u}_n^{T-t} \prod_{k=t}^{T-1} {\sigma'}_{n,k+1}|>\epsilon$. Accordingly, we can obtain the range for the recurrent weights to effectively tackle the gradient vanishing as $|u_n|> \sqrt[\leftroot{-3}\uproot{5}(T-t)]{\frac{\epsilon}{|\prod_{k=t}^{T-1} {\sigma'}_{n,k+1}|}}$. One advantage of IndRNN is that it can work robustly with ReLU as activation function. For ReLU, when the neuron is activated, its gradient is fixed to be $1$ without decay, greatly alleviating the gradient vanishing problem in the recurrent processing. Therefore, the constraint can be simplified to $|u_n|> \sqrt[\leftroot{-3}\uproot{5}(T-t)]{\epsilon}$. In other words, with the recurrent weight regulated in this range $|u_n|> \sqrt[\leftroot{-3}\uproot{5}(T-t)]{\epsilon}$, IndRNN with ReLU can avoid gradient vanishing. This is the constraint for one neuron, and for IndRNN with many neurons, this constraint can be further relaxed. For a multiple layers RNN with each layer containing multiple neurons, different neurons may learn different features including features from different ranges of memories. For IndRNN, neurons in one layer are independent from each other, the vanishing gradient of one neuron does not affect the learning of other neurons. For example, some IndRNN neurons in the first few layers may only learn the spatial information at its own time step projecting the input into a feature space, then the following layers can process such features in the temporal domain. Such neurons in the first few layers do not concern any memory and the gradient backpropagated from time is zero (gradient vanishing from the perspective of temporal gradient backpropagation for this specific neuron), but the gradients backpropagated from the deep layers at the current step still exist. Moreover, with the temporal gradient backpropagation of neurons independent in one layer, the above neurons do not affect the learning of other neurons in the same layer. Therefore, the condition of avoiding gradient vanishing shown above ($|u_n|> \sqrt[\leftroot{-3}\uproot{5}(T-t)]{\epsilon}$) can be further relaxed to $|u_n|> \sqrt[\leftroot{-3}\uproot{5}(T-t)]{\epsilon}, \exists u_n \in \mathbf{u}$, which can be easily met. Although this condition can be forcibly applied in the implementation, in practice, we found that there is no need for even such a relaxed constraint. The recurrent weights generally distribute in the whole range of memories even under no constraints as shown in the following Subsection \ref{section_vis} Fig. \ref{fig_recurrent_weight}, and highly unlikely that all the recurrent weights are trained to be close to zero with the gradient of the whole network close to zero. 

Therefore, for robust training of IndRNN without gradient exploding and vanishing, only the constraint $|u_n|\leq \sqrt[\leftroot{-3}\uproot{5}(T-t)]{\gamma}$ is needed and thus used in the experiments. Note that the regulation on the recurrent weight $u$ is different from the gradient or gradient norm clipping technique. For the gradient clipping or gradient norm clipping \cite{pascanu2013difficulty}, the calculated gradient has already exploded and is forced back to a predefined range. The gradients for the following steps may keep exploding. In this case, the gradient of the other layers relying on this neuron may not be accurate. On the contrary, the regulation proposed here essentially maintains the gradient in an appropriate range without affecting the gradient backpropagated through this neuron.

\subsection{Recurrent Weight and Memory}
\label{sec_weight_memory}
In this Subsection, the relationship between the recurrent weight and the memory keeping mechanism in IndRNN is analyzed to explicitly show the range of the recurrent weights for keeping different ranges of memories.

One of the key roles of RNNs is to keep memories of the past. That is to say, the current state at time step $T$ can be affected by the past state at time step $t$. In other words, a change of the past state ($h_t$) changes the current state ($h_T$). From the perspective of gradient backpropagation, the gradient from $h_T$ propagated to $h_t$ could still be effectively updating the weights. Assume the minimum effective gradient is $\epsilon$, and the gradient follows $|\frac{\partial {J_{n,T}}}{\partial {h}_{n,t}}|>\epsilon$. Accordingly, a range for the recurrent weight to keep memory of $T-t$ time steps can be obtained, which is $|u_n|\in (\sqrt[\leftroot{-3}\uproot{5}(T-t)]{\frac{\epsilon}{|\prod_{k=t}^{T-1} {\sigma'}_{n,k+1}|}}, +\infty)$ according to (\ref{KIRNNgradient}) (ignoring the gradient backpropagated from the objective to the hidden state at time step $T$, $\frac{\partial {J_n}}{\partial {h}_{n,T}}$). On the contrary, when only short-term memory is required, the gradient from $h_T$ propagated to $h_t$ should not be able to effectively update the weights, and the gradient follows $|\frac{\partial {J_{n,T}}}{\partial {h}_{n,t}}|\leq \epsilon$. The range of the recurrent weight can be obtained accordingly as $|u_n|\in [0, \sqrt[\leftroot{-3}\uproot{5}(T-t)]{\frac{\epsilon}{|\prod_{k=t}^{T-1} {\sigma'}_{n,k+1}|}}]$. In an extreme case, the short-term memory is no memory of the past at all, where the recurrent weight is $0$ and the gradient backpropagated from the future time steps is zero (the gradient backpropagated from the deeper layers still exists). The short-term memories can be important for the performance of the network as well. Especially for a multiple layers RNN, the short-term memories from the earlier layers may be accumulated in the following layer with long-term memory, which is also the reason that the gradient constraint on the vanishing gradient in the above subsection can be alleviated. Together with the basic requirement of avoiding the gradient exploding described above, the relationships among recurrent weight and memory through the gradient can be depicted as Table \ref{ch5_weight_memory}. With this relationship between the recurrent weight and memory, the range of memories learned by the task can be visualized through the learned recurrent weights, which is illustrated in the experiments.

Without any prior knowledge, the short-term memories and long-term memories are assumed to be both important for sequence problems. Therefore, the recurrent weight can be initialized in the range $[0, \sqrt[\leftroot{-3}\uproot{5}(T-t)]{\frac{\gamma}{|\prod_{k=t}^{T-1} {\sigma'}_{n,k+1}|}}]$, where different neurons can keep memories of different lengths. For ReLU, it can be simplified to $|u_n|\in [0, \sqrt[\leftroot{-3}\uproot{5}(T-t)]{\gamma}]$. The training of the network can further refine the network to learn required memories for the task by updating the recurrent weights. On the other hand, for tasks such as classification after reading a sequence, for the last layer, only the hidden state at the last time step is used for classification. In order to make the hidden state at the last time step rich of information as much as possible, it is better to keep the long-term memory to aggregate useful information from the beginning to the last time step. Therefore, the recurrent weights can be initialized in the range $[\sqrt[\leftroot{-3}\uproot{5}(T-t)]{\frac{\epsilon}{|\prod_{k=t}^{T-1} {\sigma'}_{n,k+1}|}}, \sqrt[\leftroot{-3}\uproot{5}(T-t)]{\frac{\gamma}{|\prod_{k=t}^{T-1} {\sigma'}_{n,k+1}|}}]$ (which is $[\sqrt[\leftroot{-3}\uproot{5}(T-t)]{\epsilon}, \sqrt[\leftroot{-3}\uproot{5}(T-t)]{\gamma}]$ for ReLU) to only keep long-term memory, which is also observed to improve performance in the experiments.

\begin{figure*}[tbp]
	\centering
	\subfloat[Sequence length 256]{
	\includegraphics[width=0.3\hsize]{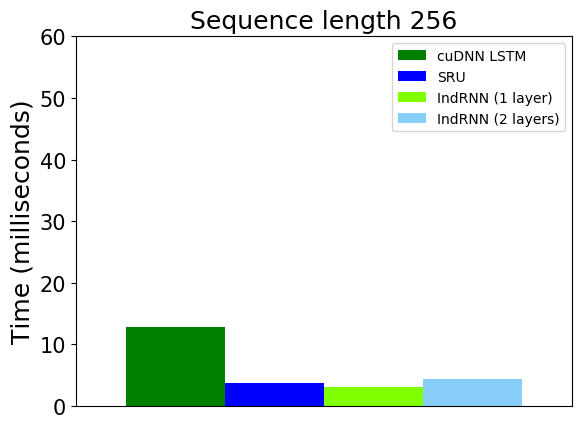}\label{img_time256}}
	\subfloat[Sequence length 512]{
	\includegraphics[width=0.3\hsize]{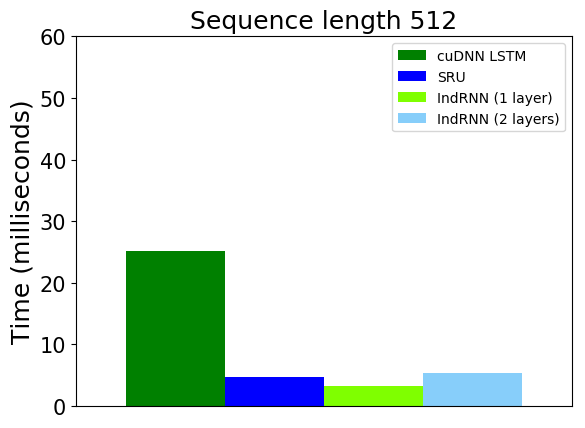}\label{img_time512}}
	\subfloat[Sequence length 1024]{
	\includegraphics[width=0.3\hsize]{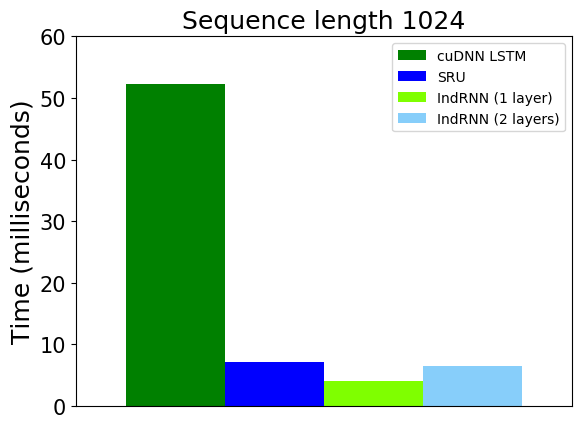}\label{img_time1024}}
	\caption{Complexity comparison in terms of time training one batch (milliseconds) for processing sequences of different lengths. Note that in our experiments the time for one forward or backward process is not very accurate due to the precision of the system clock since the time taken by the forward process is very short. Therefore, the total time of training one batch (including fetching data, forward pass, loss calculation and backward pass) is averaged over 100 batches.}
	\label{img_complexity}
\end{figure*}

\subsection{Complexity in terms of the Number of Parameters and Computation}
\label{sec_comp_neuron}
Regarding the number of parameters, for a $N$-neuron RNN network with input of dimension $M$, the number of parameters in a traditional RNN is $M\times N + N\times N + N$, while the number of parameters for LSTM is $4*(M\times N + N\times N + N)$, which is much larger than the simple RNN. By contrast, IndRNN uses a simple formulation with less parameters, $M\times N+2\times N$. Even for a two-layer IndRNN where both layers consist of $N$ neurons, the number of parameters is $M\times N+N\times N+4\times N$, which is of a similar order to the traditional RNN. Usually the number of parameters used in a network is large (e.g. much larger than $3N$), therefore, the number of parameters introduced by the recurrent weight vectors and extra bias vector ($3\times N$) is negligible compared with the recurrent weight matrix ($N\times N$) used in the traditional RNN. And obviously, IndRNN uses much less parameters than LSTM.

From the perspective of computation, the processing of the input ($\mathbf{Wx}_t+\mathbf{b}$) is independent at different timesteps and can be implemented in parallel,  which is the same for the conventional RNNs. For the processing of the recurrent input (which is usually the most time-consuming part in RNNs), IndRNN only applies one element-wise vector product operation, one adding operation and one activation calculation, involving less computation than the traditional RNNs with matrix product. Moreover, IndRNN works efficiently with ReLU, which is more efficient than other saturated functions such as the tanh and sigmoid used in the traditional RNNs. To further accelerate the computation on GPU, similarly as SRU \cite{lei2017training} we implemented a fast CUDA optimized version based on PyTorch which is made publicly available. It combines the operations of element-wise vector product, adding and activation calculation to avoid the latency in calling each function separately.

Fig. \ref{img_complexity} illustrates the computation complexity comparison in terms of computation time on different lengths of sequences including 256, 512 and 1024. The program is tested on a GPU P100 and the training time for each batch (averaged over 100 batches) including both the forward and backward time (milliseconds) are shown in the figure. In our experiments, we found that this is more accurate than using the forward and backward time separately for each batch. The one-layer IndRNN and two-layer IndRNN are evaluated and the cuDNN LSTM and SRU \cite{lei2017training} (both one layer) are used for comparison. The detailed setup is shown in Subsection \ref{subsec_adding}. It can be seen that IndRNN (both 1-layer and 2-layer) takes much less time than LSTM. To be specific, for sequences of length 256, 512 and 1024, a 1-layer IndRNN is 4.3, 7.6 and 12.9 times faster than LSTM, respectively. Even a 2-layer IndRNN is 2.9, 4.8 and 8.0 times faster than a 1-layer LSTM for sequences of length 256, 512 and 1024, respectively. It is worth noting that while the time for processing longer sequences using LSTM is dramatically increasing, the time using IndRNN only increases a very small amount. This indicates that compared with the matrix product (in the weight processing of the input), the element-wise product and addition is much faster, making IndRNN highly efficient with a comparable computation as a feedforward network. Moreover, it can also be seen that the one-layer IndRNN is faster than SRU \cite{lei2017training} and even two-layer IndRNN is faster than SRU \cite{lei2017training} for processing sequence of length 1000, which should be the case since SRU uses more computations at each time step. The memory consumptions of different RNNs also follow closely to the number of parameters and the computation, which will not be further detailed.

\begin{figure*}[tbp]
	\centering
	\subfloat[Basic IndRNN architecture]{
	\includegraphics[width=0.25\hsize]{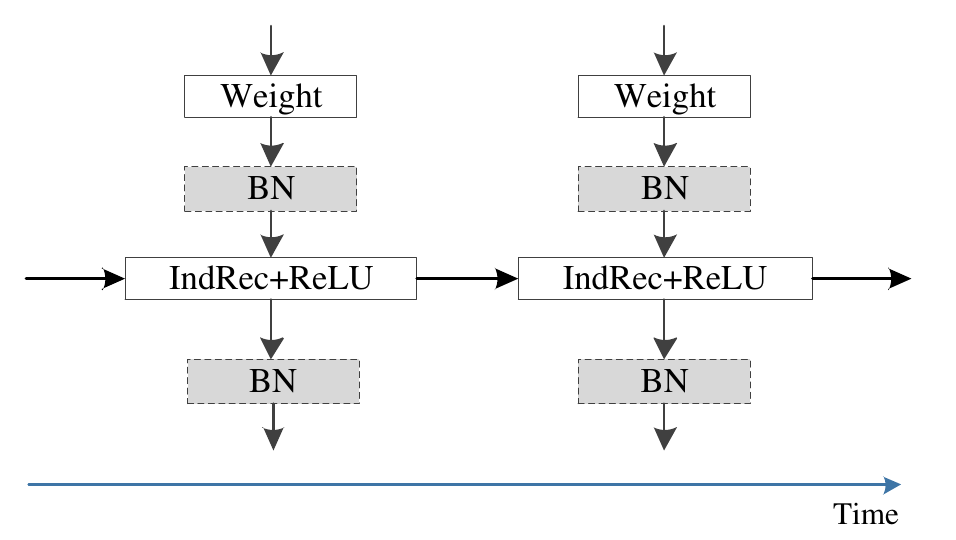}\label{img_plainIndRNN}}
	\subfloat[Residual IndRNN architecture]{
	\includegraphics[width=0.35\hsize]{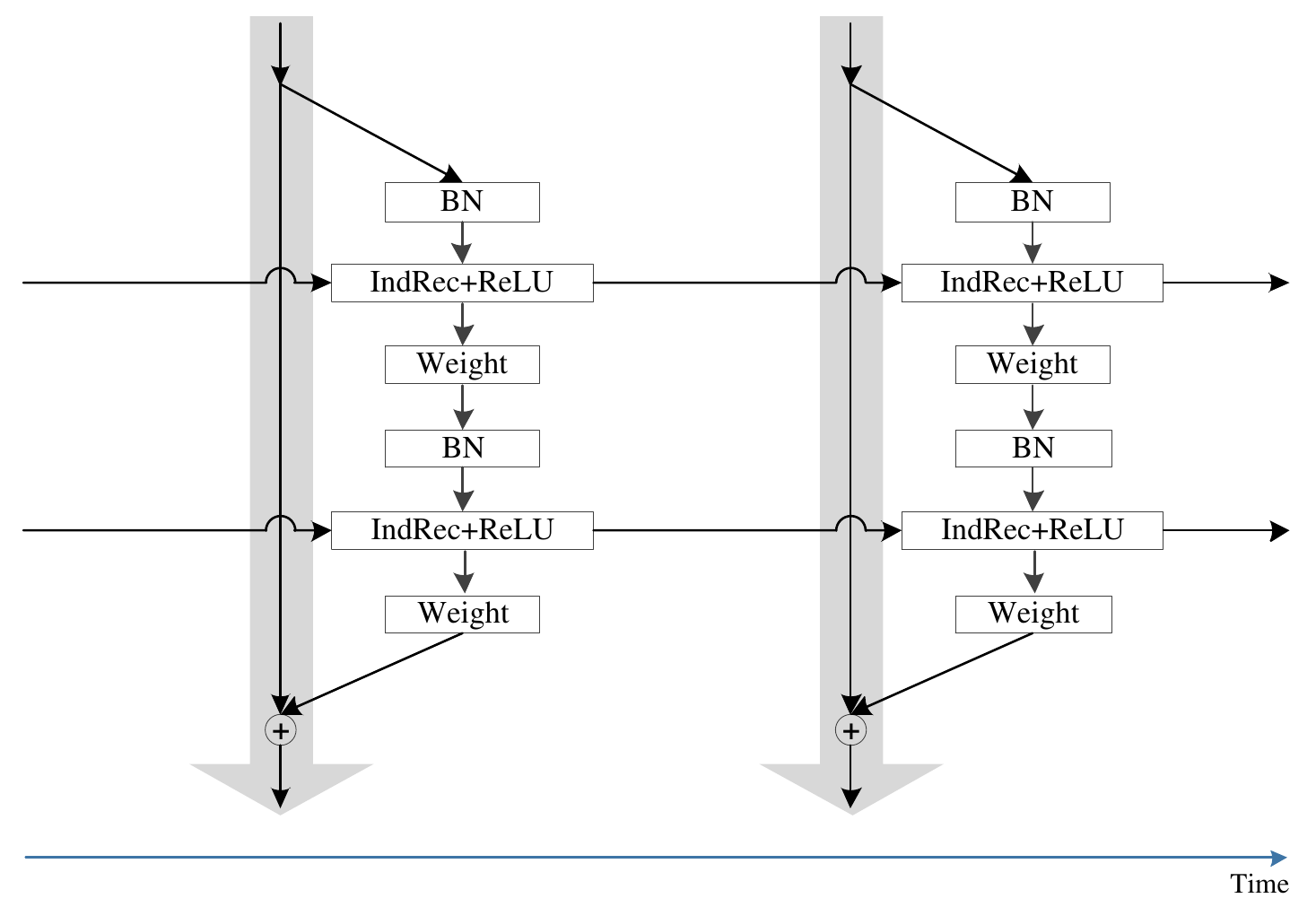}\label{img_resIndRNN}}
	\subfloat[Densely connected IndRNN architecture]{
	\includegraphics[width=0.35\hsize]{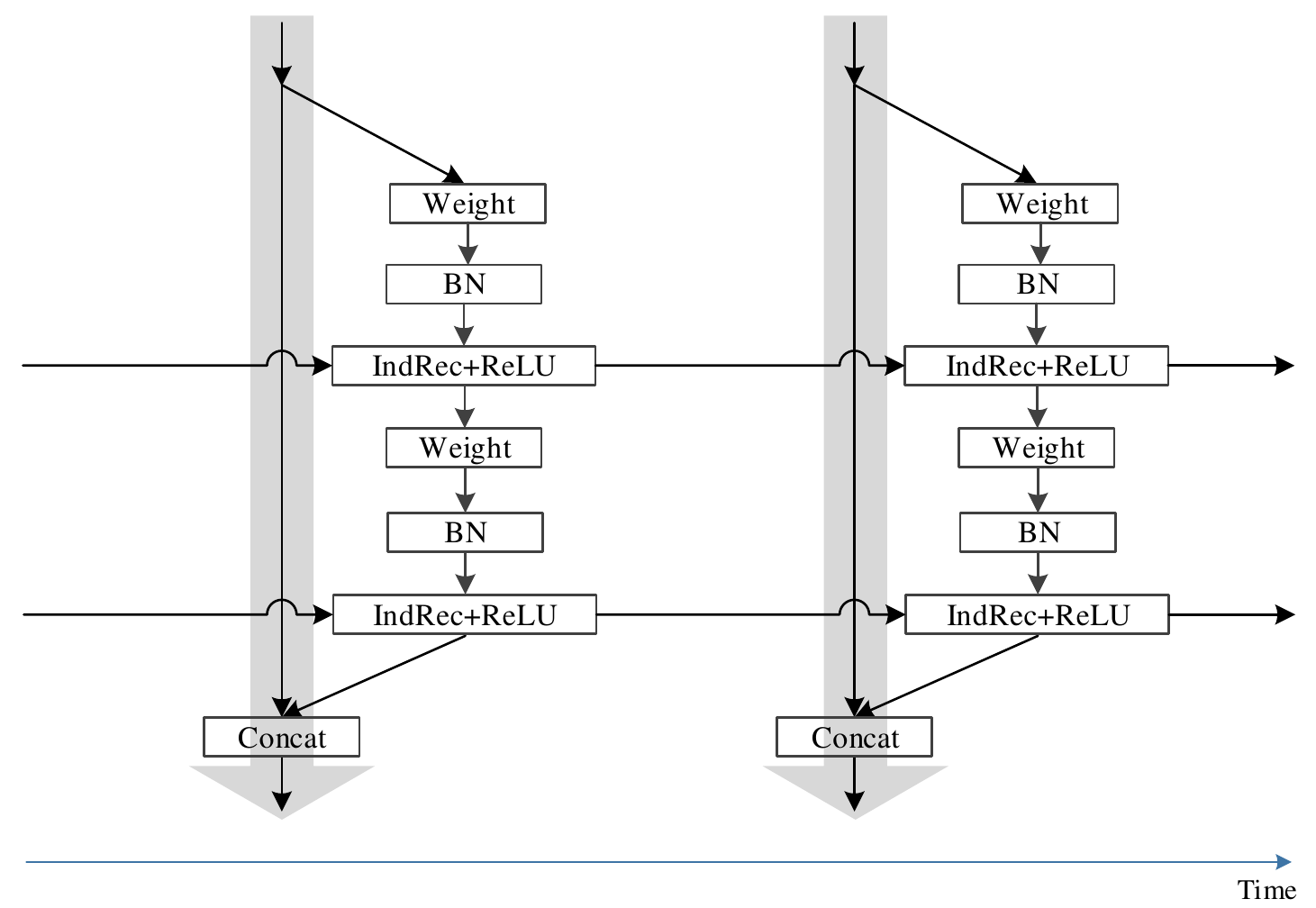}\label{img_denseIndRNN}}
	\caption{Illustration of different deep IndRNN architectures.} 
\label{rnnillustration}
\end{figure*}

\section{Deep Independently Recurrent Neural Network}
\label{sec_deepindrnn}
As described in the above Section, IndRNN can be stacked to construct multiple-layer IndRNN networks to better explore the cross-channel information. One important advantage of IndRNN is that it can work robustly with non-saturated activation functions such as ReLU, since the gradient behaviour of IndRNN can be well regulated. The non-saturated functions provide stable gradient propagation when activated (feature values larger than 0), which can greatly facilitate the development of deep networks. While in the current CNNs, non-saturated activation functions such as ReLU have been widely used and become the default setting, it is still not prevailing in the construction of RNNs due to the gradient exploding over time. Although there are some attempts on using ReLU as activation function such as IRNN \cite{le2015simple}, it is very unstable and even carefully trained with a very small learning rate ($10^{-5}$), the model may still explode. Most of the RNN models adopt gates with saturated functions to address the gradient exploding problem. To the best of our knowledge, IndRNN is the only one that stably and completely works with non-saturated activation functions such as ReLU without any gate function including saturated activation. Thus IndRNN can efficiently construct deep networks.

Several deep IndRNN architectures are proposed in this paper, including the basic stacked IndRNN, residual IndRNN (res-IndRNN) and densely connected IndRNN (dense-IndRNN). First, the building block for the basic IndRNN architecture is shown in Fig. \ref{img_plainIndRNN}, where ``Weight'' and ``IndRec+ReLU'' denote the processing of input and the recurrent process at each step with ReLU as the activation function. In addition, batch normalization, denoted as ``BN'', can also be employed in the IndRNN network before or after the activation function as shown in Fig. \ref{img_plainIndRNN}. By stacking this block, a deep basic IndRNN network can be constructed. Since the weight layer ($\mathbf{Wx}_t+\mathbf{b}$) is used to process the input, it is natural to extend it to multiple layers to deepen the processing, making the architecture flexible with different depths to process input features.

The second type of deep IndRNN architecture is the residual IndRNN shown in Fig. \ref{img_resIndRNN}, which adds an identity shortcut (skip-connection) \cite{he2016deep} to bypass the non-linear transformation of the input features. Denoting the non-linear transformation at layer $l$ and time step $t$ by $\mathcal{F}_{l,t}$, it follows $x_{l,t}=x_{l-1,t}+\mathcal{F}_{l,t}(x_{l-1,t})$, where $x_{l,t}$ is the output feature at layer $l$. Since the non-linear transformations containing IndRNN are shared over time, $\mathcal{F}_{l,t}$ can be simplified to $\mathcal{F}_{l}$ and accordingly, $x_{l,t}=x_{l-1,t}+\mathcal{F}_{l}(x_{l-1,t})$. It can be seen that the skip-connection does not affect the processing in the time dimension, but makes the deeper feature a summation of the shallower feature and a residual feature processed with $\mathcal{F}_{l}$. In this case, the gradient can be directly propagated to the previous layers through the skip-connection, greatly alleviating the gradient decay problem across layers. Therefore, substantially deeper residual IndRNN can be constructed. Fig. \ref{img_resIndRNN} shows the non-linear transformation $\mathcal{F}_{l,t}$ of type ``pre-activation'' similarly as in \cite{he2016identity}, which uses the processing of a composite function of three consecutive operations: batch normalization (BN), the recurrent processing and activation calculation (IndRec+ReLU) over time, and the weight processing (Weight), denoted by ``BN-IndRec+ReLU-Weight'' for simplicity. Also in each residual block, different numbers of ``BN-IndRec+ReLU-Weight'' operation can be used, and two operations are shown in the figure. 

The third type of deep IndRNN architecture is the densely connected IndRNN shown in Fig. \ref{img_denseIndRNN}, which, instead of using a skip-connection, uses a concatenation operation to combine all the features of the previous layers \cite{huang2017densely}. It follows $x_{l,t}=\mathcal{C}(x_{l-1,t}, \mathcal{F}_{l}(x_{l-1,t}))$, where $\mathcal{C}$ is concatenation operation. By substituting $x_{l-1,t}$ recursively, it can be easily seen that $x_{l,t}$ is a combination of all the features in the previous layers and the non-linear transformation of the features in the previous layer. This also forms an identity function among deeper layers and shallower layers, thus allowing gradients to be properly propagated. Compared with the residual IndRNN, it explicitly processes all the features from all the previous layers, encouraging the feature reuse and thus reducing the number of parameters. Different composite functions can be used as in the residual IndRNN, and Fig. \ref{img_denseIndRNN} shows an example with the composite function ``Weight-BN-IndRec+ReLU''. As the number of layers concatenated together increases, the dimension of features also increases, leading to an increased number of parameters. Therefore, transition layers as in \cite{huang2017densely} are used to separate dense layers into dense blocks to reduce the dimension of features. More details on the densely connected IndRNN architecture settings are shown in the following Subsection \ref{subsec_trainsetup} and Appendix \ref{appenx_b}.

We would like to note that while the residual connection and dense connection can also be applied to the other RNN models, they may behave better on IndRNN. As reported in \cite{veit2016residual}, residual networks behave like ensembles of relatively shallow networks. The performance of residual networks is highly dependent on the depth of the model (without residual connections) that can be constructed and effectively trained. In other words, it still depends on the gradient of the model without the residual connection. The dense connection can also be analysed similarly. Therefore, combined with the residual or dense connections, IndRNN is expected to work better than other RNN models with better gradient behaviour over layers. In all, IndRNN can construct deep RNN models and even deeper models with residual and dense connections.

\section{Experiments}
\label{sec_experiment}
In this Section, first, the capability of IndRNN in processing long sequences and constructing deep networks is verified. Then the performance of the proposed IndRNN on various tasks are evaluated.

\subsection{Training Setup}
\label{subsec_trainsetup}
ReLU was used for activation throughout the experiments. For tasks with output at every timestep such as the language modeling problem in Subsection \ref{exp_langmodel}, the recurrent weights of all layers were initialized uniformly in the range $[0, \sqrt[\leftroot{-3}\uproot{5}(T-t)]{\gamma}]$, while for tasks such as classification where only the output at the last time step in the last layer is used, the recurrent weights of the last layer were initialized uniformly in the range $[\sqrt[\leftroot{-3}\uproot{5}(T-t)]{\epsilon}, \sqrt[\leftroot{-3}\uproot{5}(T-t)]{\gamma}]$ as described in Subsection \ref{sec_weight_memory}. The recurrent weights are constrained to be in the range of $|u_n|\leq \sqrt[\leftroot{-3}\uproot{5}(T-t)]{\gamma}$ for IndRNN with ReLU as analysed in Subsection \ref{BPTT}. This is especially important in processing long sequences where a small change of recurrent weights may significantly change the gradients. In our experiments, $\gamma$ is simply set to $1.0$ for long sequences since $\sqrt[\leftroot{-3}\uproot{5}(T-t)]{\gamma}$ is already close to 1 when $T$ is very large. For short sequences such as 20 steps, the constraint can also be removed as the gradient exploding problem is not very severe in such cases. We have also conducted an ablation study on $\gamma$, which also verifies that the performance is not very sensitive to the setting of $\gamma$ as long as it is in a reasonable range (values in $[1,10]$ for example).

To accelerate training, batch normalization was used except in the simple adding problem. Moreover, for classification tasks where the whole sequence is processed for output at the final time step, the statistics used in the batch normalization layer were obtained based on the samples at all time steps, while for other tasks such as language modeling which cannot access information from future time steps, the statistics are obtained based on all the samples in each time step. When dropout is applied, the dropout mask is shared over time to avoid the clipping of long memory. Weight decay of $10^{-4}$ is used for the weight parameters (without applying to the recurrent weight and bias). All the networks were trained using the Adam optimization method \cite{kingma2014adam} with initial learning rate $2\times 10^{-4}$. The learning rate is reduced by a factor of 5 when the accuracy (or loss) on the validation data no longer improves (drops) (with patience set to 100). 

For the densely connected IndRNN, the network shape was simply following the conventional denseCNN \cite{huang2017densely}. Each dense layer (the non-linear transformation function) consists of two composite functions as shown in Fig. \ref{img_denseIndRNN} and produces k feature maps, termed as the growth rate. The first composite function is called the bottleneck layer and the number of neurons is set to four times (4*k) the growth rate. For each transition layer, the number of the neurons is set to be the half (50\%) of the input feature maps. The dense block configuration is set to (8,6,4), where in the first, second and third dense block, 8, 6 and 4 dense layers are used, respectively. This keeps a relatively similar number of neurons in each dense block. Note that it is different from the denseCNN \cite{huang2017densely} because the tasks in the following experiments do not concern pooling (which reduces the size of the features). For the whole network, one IndRNN layer with six times (6*k) the growth rate are used to process the input first before going through the following dense layers. In the following, the residual IndRNN and the densely connected IndRNN are noted as res-IndRNN and dense-IndRNN, respectively.

\begin{figure*}[tbp]
	\centering
	\subfloat[]{
	\includegraphics[width=0.35\hsize]{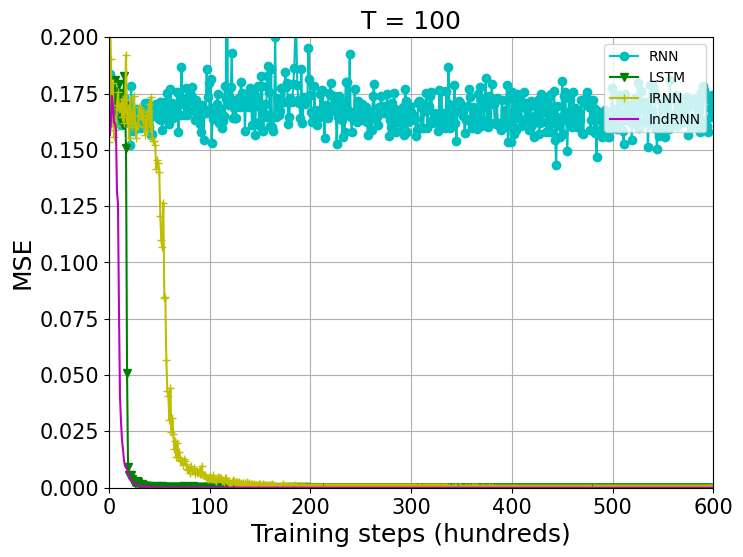}\label{adding_100}}
	\hspace{1cm}
	\subfloat[]{
	\includegraphics[width=0.35\hsize]{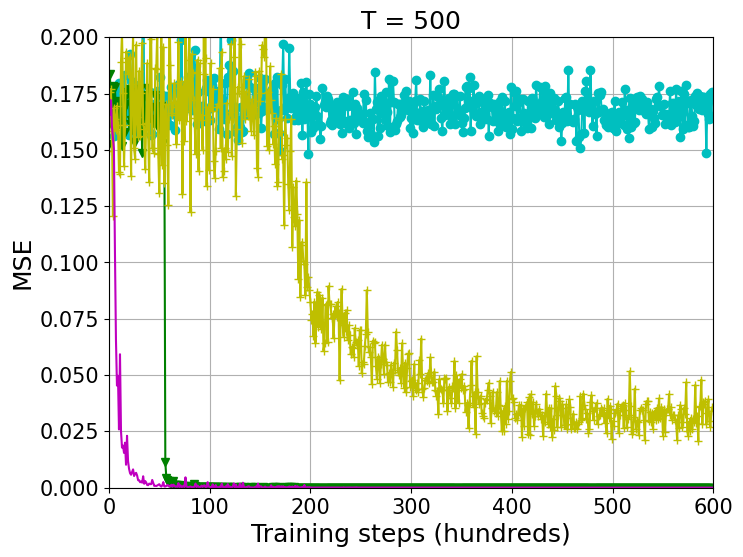}\label{adding_500}}
	
	\subfloat[]{
	\includegraphics[width=0.35\hsize]{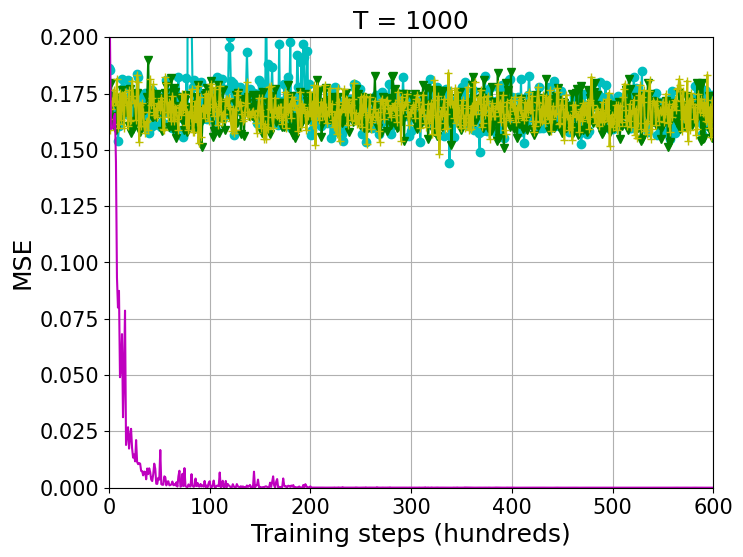}\label{adding_1000}}
	\hspace{1cm}
	\subfloat[]{
	\includegraphics[width=0.35\hsize]{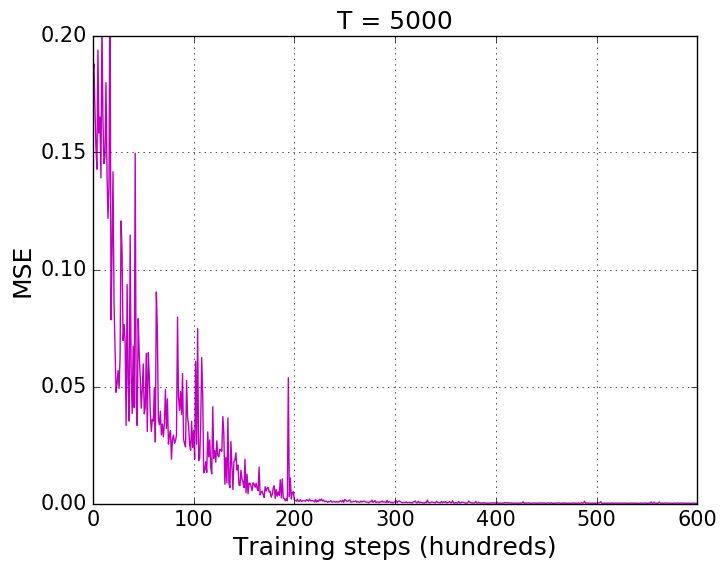}\label{adding_5000}}

	\caption{Results of the adding problem for different sequence lengths. The legends for all figures are the same and thus only shown in (a).}
	\label{adding} 
\end{figure*}

\subsection{Verification of Processing Long Sequences and Constructing Deep Networks}
\subsubsection{Adding Problem}
\label{subsec_adding}
The adding problem~\cite{hochreiter1997long, arjovsky2015unitary} is commonly used to evaluate the performance of RNN models. Two sequences of length $T$ are taken as input. The first sequence is uniformly sampled in the range $(0,1)$ while the second sequence consists of two entries being $1$ and the rest being $0$. The output is the sum of the two entries in the first sequence indicated by the two entries of $1$ in the second sequence. Three different lengths of sequences, $T=100$, $500$ and $1000$, were used for the experiments to show whether the tested models have the ability to model long-term memory.
 
The RNN models included in the experiments for comparison are the traditional RNN with tanh, LSTM, IRNN (RNN with ReLU). The proposed IndRNN was evaluated with ReLU activation function. Since GRU achieved similar performance as LSTM \cite{jozefowicz2015empirical}, it is not discussed here. RNN, LSTM, and IRNN are all one layer while the IndRNN model is two layers. $128$ hidden units were used for all the models, and the number of parameters for RNN, LSTM, and two-layer IndRNN are $16K$, $67K$ and $17K$, respectively. It can be seen that the two-layer IndRNN has a comparable number of parameters to that of the one-layer RNN, while many more parameters are needed for LSTM. As discussed in Subsection \ref{BPTT}, the recurrent weight is constrained in the range of $|u_n|\in (0, \sqrt[\leftroot{-3}\uproot{5}T]{2})$ for the IndRNN. 

Mean squared error (MSE) was used as the objective function and the Adam optimization method \cite{kingma2014adam} was used for training. The baseline performance (predicting 1 as the output regardless of the input sequence) is mean squared error of 0.167 (the variance of the sum of two independent uniform distributions). The initial learning rate was set to $2\times 10^{-3}$ for models with tanh activation and set as $2\times 10^{-4}$ for models with ReLU activations. However, as the length of the sequence increases, the IRNN model does not converge and thus a smaller initial learning rate ($10^{-5}$) was used. The learning rate was reduced by a factor of 10 every 20K training steps. The training data and testing data were all generated randomly throughout the experiments, different from \cite{arjovsky2015unitary} which only used a set of randomly pre-generated data. 

The results are shown in Fig. \ref{adding_100}, \ref{adding_500} and \ref{adding_1000}. First, for short sequences ($T=100$), most of the models (except RNN with tanh) performed well as they converged to a very small error (much smaller than the baseline). When the length of the sequences increases, the IRNN and LSTM models have difficulties in converging, and when the sequence length reaches $1000$, IRNN and LSTM cannot minimize the error any more. However, the proposed IndRNN can still converge to a small error very quickly. This indicates that the proposed IndRNN can model a longer-term memory than the traditional RNN and LSTM. 

From the figures, it can also be seen that the traditional RNN and LSTM can only keep a mid-range memory (about 500 - 1000 time steps). To evaluate the proposed IndRNN model for very long-term memory, experiments on sequences with length $5000$ were conducted where the result is shown in Fig. \ref{adding_5000}. It can be seen that IndRNN can still model it very well. Note that the noise in the result of IndRNN is because the initial learning rate ($2\times 10^{-4}$) was relatively large and once the learning rate dropped, the performance became robust. This demonstrates that IndRNN can effectively address the gradient exploding and vanishing problem over time and keep a long-term memory. 

The complexity of IndRNN was evaluated using the adding problem (with the similar configuration as above) compared with the cuDNN LSTM and SRU \cite{lei2017training} which has been reported to be much faster than LSTM. Sequences of different lengths including 256, 512 and 1024 are all tested to be consistent with \cite{lei2017training}. A GPU P100 is used as the test platform and the training time for each batch (averaged over 100 batches) including both the forward and backward time (milliseconds) are measured. The comparison is shown in Fig. \ref{img_complexity}. It can be seen that IndRNN (both 1-layer and 2-layer) takes much less time than LSTM, and reaches 12.9 times faster than LSTM when sequence length is 1024. It is also faster than SRU \cite{lei2017training}, and even a two-layer IndRNN is faster than SRU for processing sequence of length 1000. Moreover, while the time for processing longer sequences using LSTM is dramatically increasing, the time using IndRNN only increases a little bit, indicating that the recurrent connection is very efficient.

\begin{table}
\centering
\caption{Results in terms of accuracy (\%) for the sequential MNIST and permuted MNIST.} 
  \begin{tabular}{@{}l@{}cc}
  \hline
   & MNIST & pMNIST \\
  \hline
  IRNN \cite{le2015simple} & $95.0$ & $82$ \\
  uRNN \cite{arjovsky2015unitary} & $95.1$ & $91.4$\\
  RNN-path \cite{neyshabur2016path} & $96.9$ & - \\
  LSTM \cite{arjovsky2015unitary} & $98.2$ & $88$ \\ 
  LSTM+Recurrent dropout \cite{semeniuta2016recurrent} & - & $92.5$ \\ 
  LSTM+Recurrent batchnorm \cite{cooijmans2016recurrent} & - & $95.4$ \\ 
  LSTM+Zoneout \cite{krueger2016zoneout} & - & $93.1$ \\ 
  LSTM+Recurrent batchnorm+Zoneout \cite{krueger2016zoneout} & - & $95.9$ \\ 
  Transformer (reported in \cite{Wang2019RTransformerRN}) & $98.2$ & - \\ 
  R-Transformer \cite{Wang2019RTransformerRN} & $99.1$ & - \\ 
  \hline  
  \textbf{IndRNN (6 layers)} & $99.0$ & $96.0$\\
  \textbf{IndRNN (12 layers)} & $99.37$ & $96.84$\\
  \textbf{res-IndRNN (12 layers)} & $99.39$ & $97.02$\\
  \textbf{dense-IndRNN} & \textbf{99.48} & \textbf{97.20}\\
  \hline
  \end{tabular}
\label{result_mnist}
\end{table}

\subsubsection{Sequential MNIST Classification}
Sequential MNIST classification has been widely used to evaluate RNN models in capturing long dependencies. The pixels of MNIST digits \cite{lecun1998gradient} are presented sequentially to the networks and classification is performed after reading all pixels. Each digit image consists of 784 pixels ($28*28$), and to achieve a good classification accuracy, the models need to remember the patterns of such a long sequence, thus verifying the capability of the RNN models in capturing long-term dependency. To make the task even harder, the pixels are first processed with a fixed random permutation and then used for classification, known as the permuted MNIST classification. First, a six-layer IndRNN is used for test. Each layer contains 128 neurons, and batch normalization is inserted after each layer. Dropout with a dropping probability of $0.1$ is used after each layer. $5\%$ of the training data is reserved for validation. The results are shown in Table \ref{result_mnist} in comparison with the existing methods. It can be seen that IndRNN achieved better performance than the existing RNN models. 

To further verify that IndRNN can be stacked into very deep networks and still be effectively trained with a good performance, a deep IndRNN with 12 layers, a deep residual IndRNN (12 layers) and a densely connected IndRNN are also tested. For the deep IndRNN and the residual IndRNN, the settings are the same as the above 6-layer IndRNN, e.g., 128 neurons and dropout ($0.1$) applied after each layer. For the densely connected IndRNN, the growth rate is set to $16$, and dropout is applied after the input ($0.2$), each dense layer ($0.2$), each bottleneck layer ($0.1$) and each transition layer ($0.1$). The results are also shown in Table \ref{result_mnist}. Compared with the 6-layer IndRNN, deeper IndRNN (12 layers) achieves better performance, higher than $99\%$ and $97\%$ in terms of accuracy for the normal MNIST classification and permuted MNIST classification, respectively. Moreover, the residual IndRNN and the densely connected IndRNN outperforms the simple stacked IndRNN by further improving the accuracy to $99.48\%$ and $97.20\%$, respectively, validating its effectiveness.

\begin{figure*}[tbp]
	\centering
	\subfloat[]{
	\includegraphics[width=0.35\hsize]{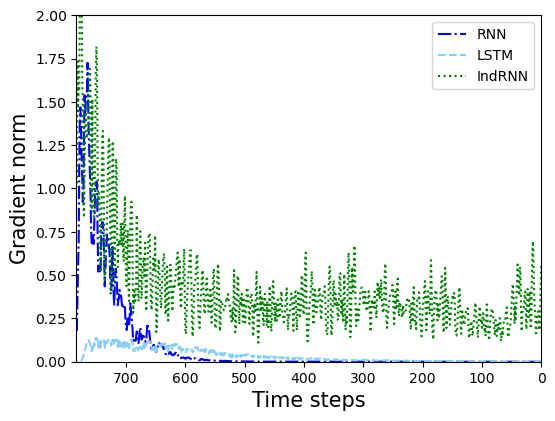}\label{grad_time}}
	\hspace{1cm}
	\subfloat[]{
	\includegraphics[width=0.35\hsize]{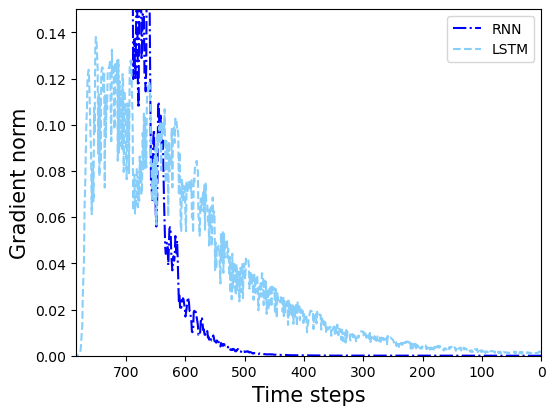}\label{grad_time_rnnlstm}}
	
	\subfloat[]{
	\includegraphics[width=0.35\hsize]{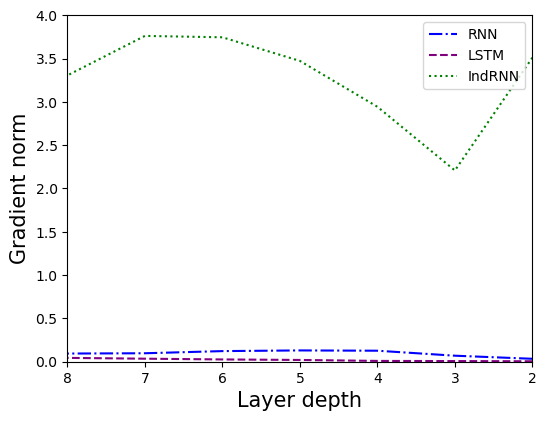}\label{grad_layer}}
	\hspace{1cm}
	\subfloat[]{
	\includegraphics[width=0.35\hsize]{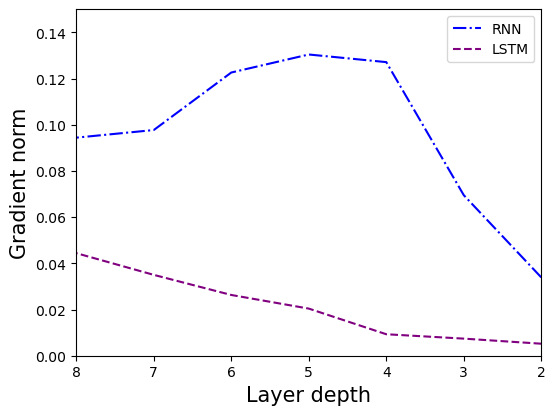}\label{grad_layer_rnnlstm}}

	\caption{Illustration of the gradient flow over time and layer depth for different RNNs. (b) and (d) are zoomed from (a) and (c), respectively, with the gradients of RNN and LSTM for better visualization and comparison. The gradient of RNN decays quickly over time and the gradient of LSTM decays quickly over layer depth.} \label{fig_gradient}
	
\end{figure*}

To experimentally compare the gradient behaviour of different models, the gradient flow over time and layer depth are illustrate in Fig. \ref{fig_gradient}. Models of 8 layers are used for different RNNs and the other settings are the same as above. The gradients are obtained under a similar training accuracy ($92\%$) for all RNNs, i.e., a similar loss and gradient backpropagated from the objective. For the gradient flow over time, the gradient backpropagated to the input is used with 784 entries. For the gradient flow over layer depth, the average gradient norm of the input processing weights, which is in the layer depth gradient backpropagation path, is used. The gradients of the first layer is omitted since it processes the pixel input with different dimensions from the other layers. The gradients are obtained as the average gradient norms over an epoch to be stable for comparison. Fig. \ref{grad_time} shows the gradient change over time and Fig. \ref{grad_time_rnnlstm} further zoomed the result of RNN and LSTM for better visualization. It can be seen that the gradient of RNN and LSTM are usually smaller than IndRNN, and the gradient decays over time. Especially for RNN, the gradient decays very quickly and close to zero in the time direction. By contrast, IndRNN maintains a relatively large gradients over time without decaying. On the other hand, as shown in Fig. \ref{grad_layer} and {\ref{grad_layer_rnnlstm}}, in the direction of layer depth, IndRNN also maintains relatively large gradients while the gradients of RNN and LSTM drops quickly. Especially for LSTM, the gradient decays severely in the layer depth direction which agrees with the analysis. It is worth noting that a LSTM of 12 layers cannot be trained to converge in our experiments mostly due to the above gradient problem. In all, the gradient of IndRNN behaves very well in both directions of time and layer depth as theoretically analysed above, and thus it is able to construct deep models and process long sequences. 

\begin{table}
\centering
\tabcolsep=12pt
\caption{Results of char-level PTB for our proposed IndRNN model in comparison with results reported in the literature, in terms of BPC.}\label{result_penntree}
\begin{threeparttable}
  \begin{tabular}{l c}
  \hline
   & Test \\
  \hline
  RNN-tanh \cite{krueger2016regularizing} & $1.55$ \\
  RNN-ReLU \cite{neyshabur2016path} &  $1.55$ \\
  RNN-TRec \cite{krueger2016regularizing} & $1.48$ \\
  RNN-path \cite{neyshabur2016path} & $1.47$ \\
  HF-MRNN \cite{mikolov2012subword} & $1.42$ \\ 
  LSTM \cite{krueger2016zoneout}  & $1.36$ \\
  LSTM+Recurrent dropout \cite{semeniuta2016recurrent} & $1.32$ \\  
  LSTM+Recurrent batchnorm \cite{cooijmans2016recurrent} & $1.32$ \\
  LSTM+Zoneout \cite{krueger2016zoneout} &  $1.27$ \\
  HyperLSTM + LN \cite{ha2016hypernetworks} & $1.25$ \\
  Hierarchical Multiscale LSTM + LN \cite{chung2016hierarchical} & $1.24$ \\
  Fast-slow LSTM \cite{mujika2017fast} & $1.19$ \\
  Neural Architecture Search \cite{zoph2016neural} & $1.21$ \\
  Transformer (reported in \cite{Wang2019RTransformerRN}) & $1.45$ \\ 
  R-Transformer \cite{Wang2019RTransformerRN} & $1.24$ \\ 
  \hline  
  \textbf{IndRNN (6 layers, 50 steps)} & $1.26$ \\
  \textbf{IndRNN (6 layers, 100 steps)} & $1.21$ \\
  \textbf{res-IndRNN (21 layers, 50 steps)} & $1.21$ \\
  \textbf{res-IndRNN (21 layers, 100 steps)} & $1.17$ \\
  \textbf{dense-IndRNN (50 steps)} & $1.19$ \\
  \textbf{dense-IndRNN (100 steps)} & \textbf{$1.16$} \\
  \hline
  \end{tabular}
\end{threeparttable}
\end{table}

\subsubsection{Char-level Penn Treebank}
\label{exp_langmodel}
The character-level and word-level language modeling tasks are also used for evaluation. 
In this subsection, we first evaluate the performance using the character-level Penn Treebank (PTB-c) dataset. The test setting is similar to \cite{cooijmans2016recurrent}. Different network architectures are also tested, including a six-layer IndRNN, a 21-layer residual IndRNN (to demonstrate that the IndRNN network can be very deep with the residual connections) and a densely connected IndRNN. For the six-layer IndRNN and the 21-layer residual IndRNN, $2000$ hidden neurons are used and dropout with dropping probabilities of $0.25$ and $0.3$ were used, respectively. While for the densely connected IndRNN, the growth rate is set to 256 and 600 hidden neurons are used for the embedding layer. Dropout is applied after each dense layer ($0.5$), each bottleneck layer ($0.1$) and each transition layer ($0.4$). Dropout of $0.3$ is used at the last transition layer, which is connected to the output layer. The batch size is set to $128$ and the sequence length $T=50$ and $T=100$ were both tested in training and testing.

Table \ref{result_penntree} shows the performance of the proposed IndRNN models in comparison with the existing methods, in terms of bits per character metric (BPC). Compared with the traditional RNN and LSTM models, it can be seen that the proposed IndRNN model achieved better performance. In addition, most IndRNN models in Table~\ref{result_penntree} except the one with 6 layers and 50 steps outperformed the popular language model Transformer. By comparing the performance of the proposed IndRNN model with different depth, it can be seen that a deeper residual IndRNN or a densely connected IndRNN further improves the performance over the relatively shallow (6 layers) model. Moreover, an improvement can be further achieved with longer temporal dependencies (from time step 50 to 100), indicating that the proposed IndRNN can effectively learn long-term patterns.

\subsubsection{Word-level Penn Treebank}
In this subsection, the performance of the proposed IndRNN on the word-level Penn Treebank dataset is evaluated. The test setting is similar to \cite{krueger2016zoneout}. A 12-layer residual IndRNN with 2000 neurons in each layer and a densely connected IndRNN with the growth rate of 256 were used for test. An embedding layer with $600$ neurons is used. The weight tying \cite{inan2016tying,press2016using} of the input embedding and the final output weight is adopted and accordingly the last transition layer contains $600$ neurons, which processes the IndRNN features from $2000$ to $600$. For the residual IndRNN, dropout with a dropping probability of $0.45$ was used among IndRNN layers while $0.5$ was used for the last IndRNN layer. For the densely connected IndRNN, dropout is applied after each dense layer ($0.5$), each bottleneck layer ($0.1$) and each transition layer ($0.4$). Dropout of $0.65$ is used after the embedding layer and the last transition layer, and dropout ($0.2$) on the embedding weight as in\cite{merity2018regularizing} is also applied for both IndRNN networks. The batch size is set to $128$. Sequence length $T=50$ was used in training and testing.

The results are shown in Table \ref{result_wPTB} in comparison with the existing methods. It can be seen that the proposed IndRNN model achieved better performance than the traditional RNN and LSTM models including the AWD-LSTM \cite{merity2018regularizing}, which is heavily optimized with weight drop, activation regularization, etc. The results can be further enhanced by post-processing with techniques such as neural cache models \cite{grave2017improving}, mixture of ssoftmaxes (MOS)\cite{Yang2017BreakingTS} and dynamic evaluation \cite{krause2018dynamic}. Here, dynamic evaluation\cite{krause2018dynamic} is used as an example and the others are not further discussed. The results are shown in the end of Table \ref{result_wPTB} with a perplexity of 50.97, which is also better than the AWD-LSTM \cite{merity2018regularizing} with the dynamic evaluation. Notice that both res-IndRNN and dense-IndRNN models outperformed the Transformer as shown in Table~\ref{result_wPTB}.

\begin{table}
\centering
\tabcolsep=12pt
\caption{Results of word-level PTB for our proposed IndRNN model in comparison with results reported in the literature, in terms of perplexity.}\label{result_wPTB}
  \begin{tabular}{l c}
  \hline
   & Test \\
  \hline
  RNN-LDA + KN-5 + cache \cite{mikolov2012context}  & $92.0$ \\
  Deep RNN \cite{pascanu2013construct}  & $107.5$ \\
  CharCNN \cite{kim2016character}  & $78.9$ \\
  LSTM \cite{krueger2016zoneout}  & $114.5$ \\
  LSTM+Recurrent dropout \cite{semeniuta2016recurrent} & $87.0$ \\  
  LSTM+Zoneout \cite{krueger2016zoneout} &  $77.4$ \\
  LSTM+Variational Dropout \cite{gal2016theoretically} & $73.4$ \\
  Pointer Sentinel LSTM \cite{merity2016pointer} & $70.9$ \\
  RHN \cite{zilly2016recurrent} & $65.4$ \\
  Neural Architecture Search \cite{zoph2016neural} & $62.4$ \\
  AWD-LSTM \cite{merity2018regularizing} & $58.8$ \\
  AWD-LSTM+Finetue \cite{merity2018regularizing} & $57.3$ \\
  AWD-LSTM+Finetue \cite{merity2018regularizing}+dynamic eval\cite{krause2018dynamic} & $51.1$ \\
  SRU (\cite{lei2017training}) & $60.3$  \\ 
  Transformer (reported in \cite{Wang2019RTransformerRN}) & $122.37$  \\ 
  R-Transformer \cite{Wang2019RTransformerRN} & $84.38$  \\ 
  \hline  
  \textbf{res-IndRNN (12 layers)} & $58.99$ \\
  \textbf{dense-IndRNN} & $55.24$ \\
  \textbf{dense-IndRNN+dynamic eval\cite{krause2018dynamic}} & \textbf{$49.95$} \\
  \hline
  \end{tabular}
\end{table}

\begin{table}
\caption{Result comparison of the RNN based skeleton action recognition methods on NTU RGB+D dataset.}
\label{result_ablation_ntu}
\begin{center}
  \begin{tabular}{l c c}
  \hline
  Method & CS & CV \\
  \hline
  1 Layer RNN (reported in \cite{shahroudy2016ntu}) & 56.02\% & 60.24\%  \\
  2 Layer RNN (reported in \cite{shahroudy2016ntu}) & 56.29\% & 64.09\%  \\
  1 Layer LSTM (reported in \cite{shahroudy2016ntu}) & 59.14\% & 66.81\%  \\
  2 Layer LSTM (reported in \cite{shahroudy2016ntu}) & 60.09\% & 67.29\%  \\
  \hline
  1 Layer RNN & 57.84\% & 64.28\%  \\
  2 Layer RNN & 62.17\% & 71.00\%  \\
  4 Layer RNN & 61.78\% & 68.28\%  \\
  6 Layer RNN & 58.23\% & 63.49\%  \\
  \hline
  1 Layer LSTM & 66.01\% & 73.87\%  \\
  2 Layer LSTM & 67.93\% & 77.55\%  \\
  4 Layer LSTM & 66.02\% & 69.12\%  \\
  6 Layer LSTM & 58.50\% & 67.42\%  \\
  \hline
  \textbf{IndRNN-tanh (4 layers)} & 76.56\% & 85.32\%  \\
  \textbf{IndRNN-tanh (6 layers)} & 80.36\% & 88.61\%  \\
  
  \textbf{IndRNN (4 layers)} & 77.23\% & 88.30\%  \\
  \textbf{IndRNN (6 layers)} & 80.44\% & 89.45\%  \\
  \textbf{res-IndRNN (100 layers)} & 83.26\% & 92.00\%  \\
  \textbf{dense-IndRNN} & \textbf{83.38\%} & \textbf{91.81\%}  \\
  \hline
  \end{tabular}
\end{center}
\end{table}

\begin{table}
\caption{Results of all skeleton based methods on NTU RGB+D dataset.}
\label{result_ntu}
\begin{center}
  \begin{tabular}{l c c}
  \hline
  Method & CS & CV \\
  \hline
  SkeletonNet(CNN) \cite{ke2017skeletonnet} & 75.94\% & 81.16\%  \\
  JDM+CNN \cite{li2017joint} & 76.20\% & 82.30\%  \\
  Clips+CNN+MTLN \cite{ke2017new} & 79.57\% & 84.83\%  \\
  Enhanced Visualization+CNN \cite{liu2017enhanced} & 80.03\% & 87.21\%  \\
  HCN \cite{ijcai2018-109} & 86.5\% & 91.1\%  \\
  TCN + TTN\cite{Lohit2019Temporal} & 77.55\% & 84.25\%  \\
  \hline
  STGCN\cite{yan2018spatial} & 81.5\% & 88.3\%  \\
  PB-GCN\cite{thakkar2018part} & \textbf{87.5\%} & 93.2\%  \\
  \hline
  1 Layer PLSTM \cite{shahroudy2016ntu} & 62.05\% & 69.40\%  \\
  2 Layer PLSTM \cite{shahroudy2016ntu} & 62.93\% & 70.27\%  \\
  ST-LSTM + Trust Gate \cite{Liu2018SkeletonBasedAR} & 69.2\% & 77.7\%  \\
  JL\_d+RNN \cite{zhang2017geometric} & 70.26\% & 82.39\%  \\
  STA-LSTM \cite{song2017end} & 73.40\% & 81.20\%  \\
  Pose conditioned STA-LSTM\cite{baradel2017pose} & 77.10\% & 84.50\%  \\
  R-Transformer \cite{Wang2019RTransformerRN} & 75.69\% & 81.56\% \\
  \hline
  \textbf{dense-IndRNN-aug} & 86.70\% & \textbf{93.97\%}  \\
  \hline
  \end{tabular}
\end{center}
\end{table}

\subsection{Skeleton based Action Recognition}
\subsubsection{NTU RGB+D dataset}
In this subsection, the skeleton based action recognition \cite{shahroudy2016ntu, Shahroudy2018DeepMF} is used to evaluate the performance of the proposed IndRNN. The widely used NTU RGB+D dataset \cite{shahroudy2016ntu}. The NTU RGB+D dataset contains $56880$ sequences (over 4 million frames) of 60 action classes, including Cross-Subject (CS) (40320 and 16560 samples for training and testing, respectively) and Cross-View (CV) (37920 and 18960 samples for training and testing, respectively) evaluation protocols \cite{shahroudy2016ntu}. In each evaluation protocol, 5\% of the training data (randomly selected) was used for evaluation as suggested in \cite{shahroudy2016ntu}. The joint coordinates of two subject skeletons were used as input. If only one is present, the second was set as zero. For this dataset, when multiple skeletons are present in the scene, the skeleton identity captured by the Kinect sensor may be changed over time. Therefore, an alignment process was first applied to keep the same skeleton saved in the same data array over time. 20 frames were sampled from each instance as one input in the same way as in \cite{liu2016spatio} and batch size was set to 128. 

First, IndRNN with different settings are evaluated first, including different IndRNN architectures (plain IndRNN and dense IndRNN), different depths (4 layers and 6 layers) and different activation functions (ReLU and Tanh). To be specific, a four-layer IndRNN and a six-layer IndRNN with $512$ hidden neurons were both tested to demonstrate the effectiveness of the proposed IndRNN with a plain IndRNN architecture. ReLU and tanh are both evaluated to show the performance of IndRNN under different activation functions. Then a densely connected IndRNN is used to further validate the effectiveness of a deep architecture. The growth rate is set to 48. For the four-layer IndRNN and six-layer IndRNN, dropout \cite{gal2016theoretically} was applied after each IndRNN layer with a dropping probability of $0.4$ and $0.3$ for models with ReLU and tanh, respectively, for both CS and CV settings. For the densely connected IndRNN, dropout is applied after the input processing layer ($0.5$), each dense layer ($0.5$), each bottleneck layer ($0.1$) and each transition layer ($0.3$). For comparison, we also included the deep RNN and LSTM of different layers to show the difference in constructing deep RNNs. The settings of the RNN and LSTM models are similar to \cite{shahroudy2016ntu} with 128 neurons for LSTM and 512 neurons for RNN (both with dropout 0.5). 1 layer, 2 layer, 4 layers and 6 layers are used for evaluation. 

The results of different settings are shown in Table \ref{result_ablation_ntu}. By comparing the performance of RNN, LSTM and plain IndRNN, it can be seen that the proposed IndRNN greatly improves the performance over the conventional RNN models on the same task. For CS, LSTM of 2 layers can only achieve accuracies of $67.93\%$ while a 4-layer IndRNN already achieved $78.58\%$, improving performance by over $10\%$. For CV, LSTM of 2 layers only achieved accuracies of $77.55\%$ while 4-layer IndRNN already achieved $83.75\%$, improving performance by over $6\%$. The results of RNN and LSTM in our experiments are higher than those reported in \cite{shahroudy2016ntu} and here we use the better results for comparison. On the other hand, by comparing the performance of RNN, LSTM and IndRNN with different layers, it can be seen that the performance of RNN and LSTM cannot be further improved by simply increasing the number of layers, which is the same as demonstrated in \cite{liu2016spatio,shahroudy2016ntu}. On the contrary, by increasing the 4-layer IndRNN to a 6-layer IndRNN, the performance is further improved to $81.80\%$ and $87.97\%$ for CS and CV, respectively. Moreover, with a deep densely connected IndRNN, the performance is further improved to $84.88\%$ and $90.34\%$ for CS and CV, respectively, which is significantly better than the conventional RNN and LSTM methods. From the perspective of different activation functions, it can be seen that IndRNN with tanh also achieves much better performance than the conventional LSTM, although slightly worse than IndRNN with ReLU. Considering the wide use and spread of the non-saturated activation function such as ReLU, we focus on IndRNN with ReLU. More importantly, ReLU does not introduce any weight decay when activated, making the gradient vanishing problem greatly alleviated. Therefore, we only presented the results of IndRNN with tanh in this experiment and not further span on it.

To further demonstrate that IndRNN can construct very deep networks, a 100 layers IndRNN with residual connections is further tested. The setting is the same as above, only the learning rate is set to $6*10^{-4}$ to reduce overfitting. The results are also shown in Table \ref{result_ablation_ntu}. It can be seen that a deeper IndRNN still can be trained robustly with improved performance. We further conducted an ablation study on the $\gamma$, the largest gradient magnitude to avoid gradient exploding, which is the only hyperparameter introduced by IndRNN. It also controls the range of the recurrent weights by $|u_n|\leq \sqrt[\leftroot{-3}\uproot{5}(T-t)]{\gamma}$. Experiments with different $\gamma$s are conducted on the CS setting with a 6-layer IndRNN and results are shown in Table \ref{result_ablation_gamma}. It can be seen that the results are relatively close, indicating that IndRNN is not very sensitive to the choice of $\gamma$ when processing short sequences. Moreover, $\gamma$ can be simply set to $1$, which shows a good performance and avoids the gradient vanishing and exploding as described in Subsection \ref{BPTT}. Accordingly, the recurrent weights are constrained to be in the range of $[-1, 1]$.

\begin{table}
\caption{Ablation study on the $\gamma$ (recurrent weight constraint).}
\label{result_ablation_gamma}
\begin{center}
  \begin{tabular}{l c c c c}
  \hline
  $\gamma$ & $1$  & $2$ & $10$ & No Constraint \\
           & $80.82\%$ & $80.68\%$ & $80.43\%$ & $80.24\%$\\
  \hline
  \end{tabular}
\end{center}
\end{table}

The final results on the NTU RGB+D dataset are shown in Table \ref{result_ntu} including comparisons with the existing methods. Considering that feature augmentation including the geometric features \cite{zhang2017geometric} and temporal feature augmentation \cite{thakkar2018part} have shown to be useful, feature augmentation is further adopted on top of the densely connected IndRNN. In addition, we also applied the data augmentation in the time dimension where sequences of length 10-30 are randomly sampled as input. The result is also shown in Table \ref{result_ntu}. It can be seen that the performance is further improved and better than the RNN based methods even equipped with enhancing techniques such as attention. 

It is noted that the performance is slightly worse than the GCN based models \cite{thakkar2018part}, which applied graph convolution to process the input spatially. However, the performance of IndRNN is only achieved with the coordinates as input without any spatial processing techniques. It should be pointed out that GCN can be used together with IndRNN to take the advantages of both networks, for instance, skeleton features instead of simple coordinates can be extracted using a GCN in a local spatial/temporal window and used as input to an IndRNN. This will be studied in the future.

\begin{table}
\caption{Results of all skeleton based methods on NTU RGB+D 120 dataset.}
\label{result_ntu120}
\begin{center}
  \begin{tabular}{l c c}
  \hline
  Method & CS & CV \\
  \hline
  Part-aware LSTM \cite{shahroudy2016ntu} (reported in \cite{Liu2019NTUR1}) & 25.5\% & 26.3\%  \\  
  Soft RNN \cite{Hu2019EarlyAP} (reported in \cite{Liu2019NTUR1}) & 36.3\% & 44.9\%  \\
  ST-LSTM \cite{liu2016spatio} & 55.7\% & 57.9\%  \\
  GCA-LSTM \cite{Liu2017GlobalCA} & 58.3\% & 59.2\%  \\
  Two-Stream Attention LSTM \cite{Liu2018SkeletonBasedHA} & 61.2\% & 63.3\%  \\
  Body Pose Evolution Map \cite{Liu2018RecognizingHA} & 64.6\% & 66.9\%  \\  
  SkeleMotion \cite{Caetano2019SkeleMotionAN} & 66.9\% & 67.7\%  \\
  \hline
  \textbf{IndRNN (4 layers)} & 70.01\% & 72.05\%  \\
  \textbf{IndRNN (6 layers)} & 72.39\% & 75.04\%  \\
  \textbf{dense-IndRNN} & 74.62\% & 77.37\%  \\
  \textbf{dense-IndRNN-aug} & \textbf{76.55\%} & \textbf{79.18\%}  \\
  \hline
  \end{tabular}
\end{center}
\end{table}

\begin{figure*}[tbp]
	\centering
	\subfloat[]{
	\includegraphics[width=0.3\hsize]{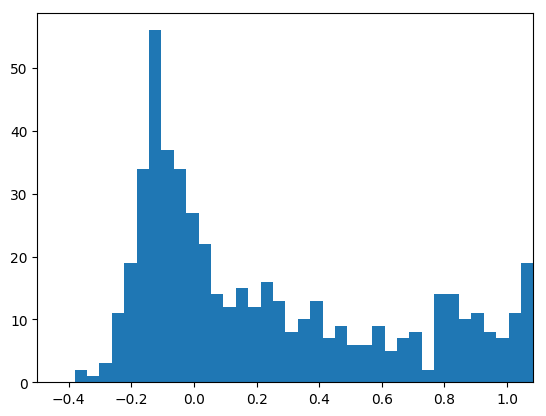}\label{fig_recurrent_weight_0}}
	\subfloat[]{
	\includegraphics[width=0.3\hsize]{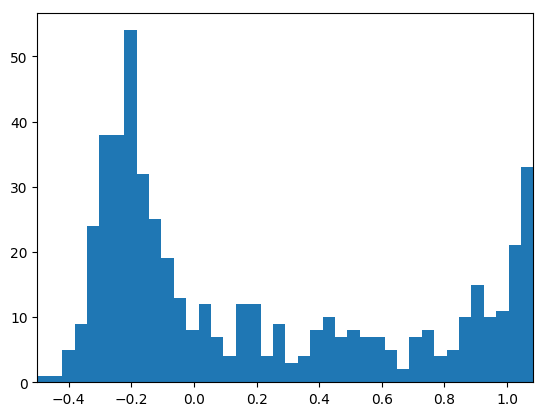}\label{fig_recurrent_weight_1}}	
	\subfloat[]{
	\includegraphics[width=0.3\hsize]{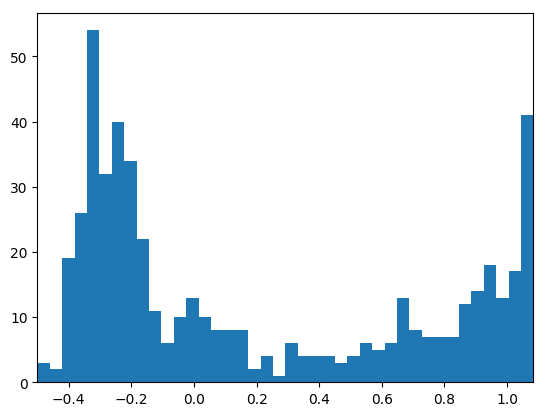}\label{fig_recurrent_weight_2}}
	
	\subfloat[]{
	\includegraphics[width=0.3\hsize]{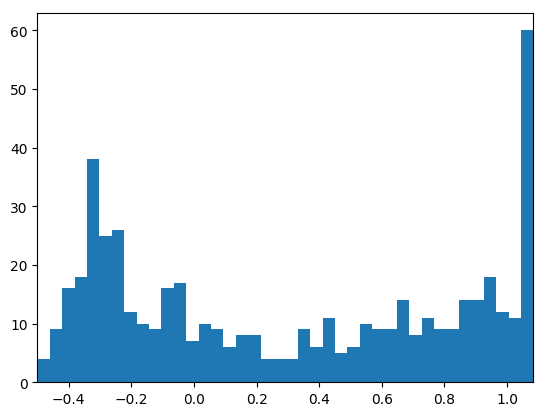}\label{fig_recurrent_weight_3}}	
	\subfloat[]{
	\includegraphics[width=0.3\hsize]{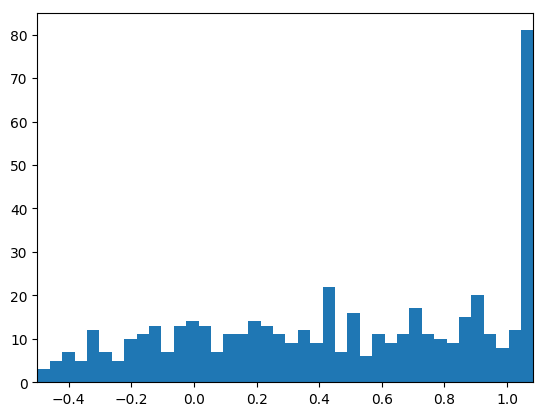}\label{fig_recurrent_weight_4}}
	\subfloat[]{
	\includegraphics[width=0.3\hsize]{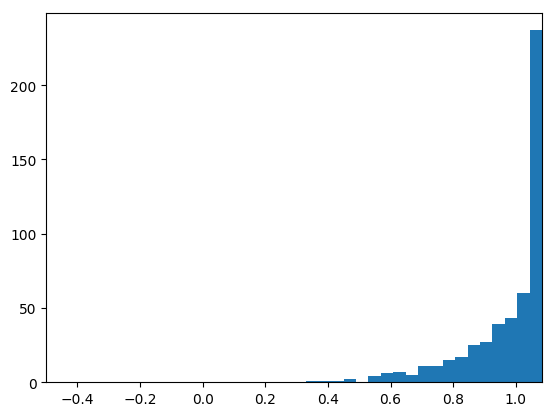}\label{fig_recurrent_weight_5}}
	\caption{Histograms of the learned recurrent weights for different layers, where (a)-(f) correspond to the 1st-6th layers. x axis and y axis represent the value of the recurrent weight and the frequency.} 
	\label{fig_recurrent_weight}
\end{figure*}

\subsubsection{NTU RGB+D 120 dataset}
The new NTU RGB+D 120 dataset \cite{Liu2019NTUR1}, enhanced on the NTU RGB+D dataset is also adopted for evaluation. It contains $114480$ video samples (over 8 million frames) from 120 different action classes, by adding another 57600 samples and another 60 classes from the NTU RGB+D dataset. Both the CS and CV evaluation protocols from the NTU RGB+D dataset are used. The other settings include the training and testing setups are the same as the one used in the above NTU RGB+D dataset.

The results on the NTU RGB+D 120 dataset are shown in Table \ref{result_ntu120} with comparisons against the existing methods. Similar to the results on the NTU RGB+D dataset, the proposed IndRNN achieves much better performance against the conventional RNN/LSTM based methods. It reaches 74.60\% and 77.37\% in terms of accuracy for the CS and CV settings, respectively, which is much better than the 61.2\% and 63.3\% for attention enhanced LSTM methods and the state-of-the-art methods using CNN with accuracy of 66.9\% and 67.7\%.

\subsubsection{Visualization of the Recurrent Weight and Memory}
\label{section_vis}
With the mapping between the recurrent weight and memory shown in subsection \ref{sec_weight_memory}, the memory learned by the network for the task can be understood by visualizing the recurrent weights. Fig. \ref{fig_recurrent_weight} shows the histograms of the learned recurrent weights (the above six-layer IndRNN obtained under the CS setting for the NTU RGB+D dataset). It can be seen that for the first 5 layers, the learned recurrent weights severely deviate from the uniformly initialized recurrent weights. Especially for the higher layers such as layer 4 and layer 5, the recurrent weights are mostly around $1$, learning the long-term memory. On the contrary, for shallower layers such as layer 0 and layer 1, in addition to the large recurrent weights around $1$, there is also a group of weights being close to $0$, learning the very short-term memory. This makes sense since the shallow layers process the very basic features in a short period while as the layers increase, high-level features in a longer period are processed. Surprisingly, the number of recurrent weights around $0.5$ is small indicating that only a small number of neurons keep mid-range memory. Another interesting fact is that although the recurrent weights are initialized to be positive, part of the recurrent weights are learned to be negative. Note that for keeping different ranges of memories, only the absolute value of the recurrent weights are relevant and thus the negative recurrent weights can still keep different ranges of memories. However, negative recurrent weights may cause oscillation as shown in the gradient backpropagation process. It is hard to intuitively understand how the negative recurrent weights work in the network yet. For the last layer, it can be seen that most of the recurrent weights stay closely to $1$ to keep long-term memory, which agrees with the previous assumption shown at the end of Section \ref{sec_weight_memory}.

\section{Conclusion}
\label{sec_conclusion}
In this paper, we proposed an independently recurrent neural network (IndRNN) with a new recurrent connection using the Hadamard product, making neurons in one layer independent of each other. It effectively solves the gradient vanishing and exploding problems by regulating the recurrent weights, which makes it able to be efficiently process very long sequences. Different deep IndRNN architectures, including the basic IndRNN, residual IndRNN and densely connected IndRNN, have been proposed, which can be much deeper than the traditional RNNs. With the gradient better regulated in IndRNN, non-saturated activation functions such as ReLU can be used and trained very robustly. Moreover, with the neurons in one layer independent from each other, each neuron can be better interpreted without effects from others. Also, the relationship between the memory and the recurrent weight has been established through gradient backpropagation, and the learned memories can be understood by investigating the recurrent weights. Experiments on multiple fundamental tasks have verified the advantages of the proposed IndRNN over existing RNN models. 

\bibliographystyle{IEEEtran}
\bibliography{reRNNreference}

\begin{IEEEbiography}[{\includegraphics[width=1in,height=1.25in,clip,keepaspectratio]{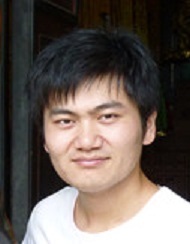}}]{Shuai Li}
is currently with the School of Control Science and Engineering, Shandong University (SDU), China, as a Professor and QiLu Young Scholar. He was with the School of Information and Communication Engineering, University of Electronic Science and Technology of China (UESTC), China, as an Associate Professor from 2018-2020. He received his Ph.D. degree from the University of Wollongong, Australia, in 2018. His research interests include image/video coding, 3D video processing and computer vision. He was a co-recipient of two best paper awards at the IEEE BMSB 2014 and IIH-MSP 2013, respectively.
\end{IEEEbiography}

\begin{IEEEbiography}[{\includegraphics[width=1in,height=1.25in,clip,keepaspectratio]{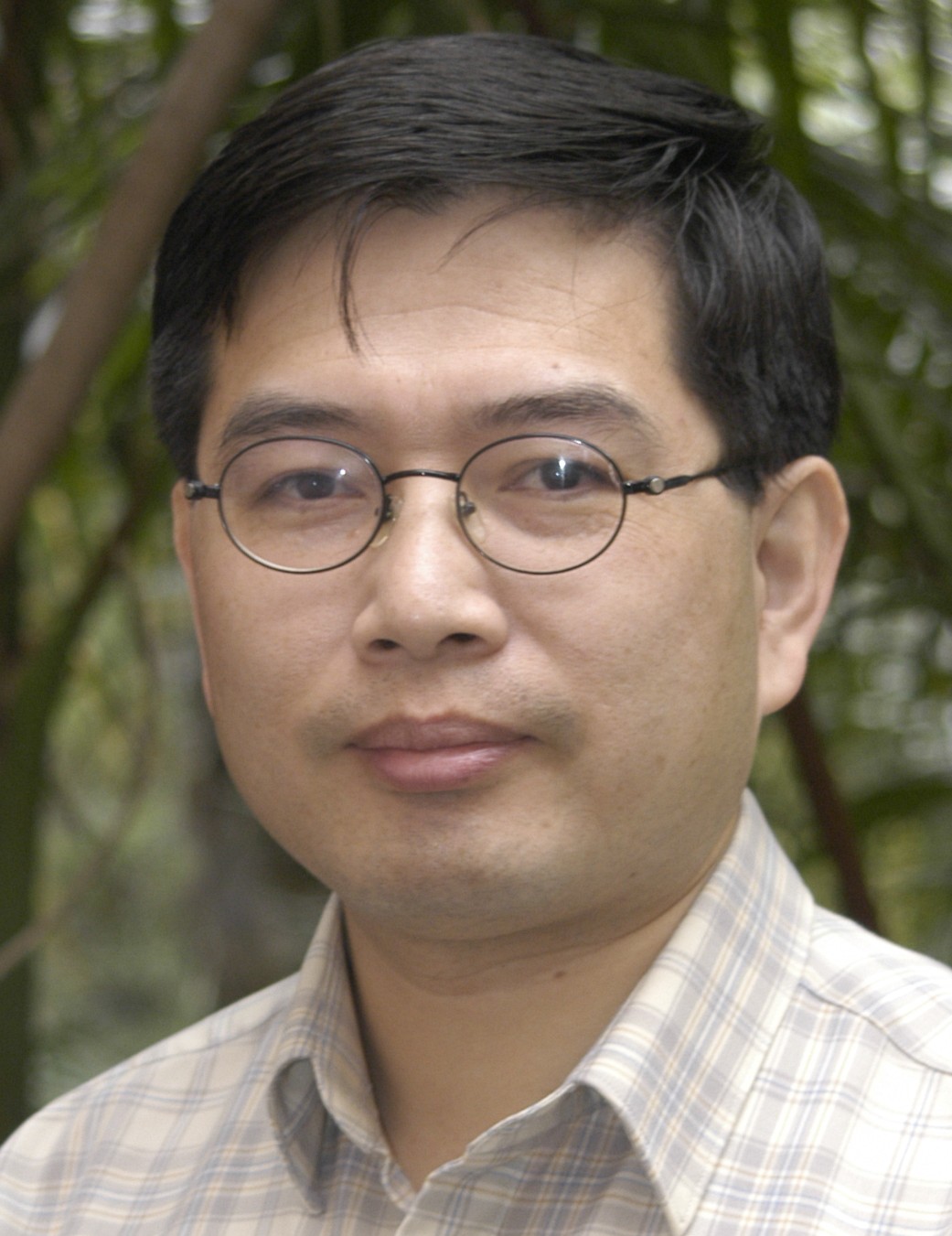}}]{Wanqing Li} (M'97-SM'05) received his Ph.D. in electronic engineering from The University of Western Australia. He was a Principal Researcher (98-03) at Motorola Research Lab and a visiting researcher (08, 10 and 13) at Microsoft Research US. He is currently an Associate Professor and Co-Director of Advanced Multimedia Research Lab (AMRL) of UOW, Australia. His research areas are machine learning, 3D computer vision, 3D multimedia signal processing and medical image analysis. Dr. Li currently serves as an Associate Editor for \textsc{IEEE Transactions on Circuits and Systems for Video Technology} and \textsc{IEEE Transactions on Multimedia}. He was an Associator for \textsc{Journal of Visual Communication and Image Representation}.
\end{IEEEbiography}

\begin{IEEEbiography}[{\includegraphics[width=1in,height=1.25in,clip,keepaspectratio]{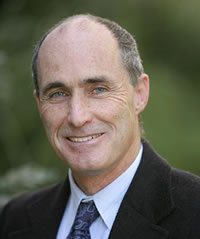}}]{Chris Cook}
is an electrical engineer with a BSc and BE from the University of Adelaide, and he received his PhD from the University of New South Wales in 1976. He is a Fellow of the Institute of Engineers Australia and a Chartered Engineer. He has worked for Marconi Avionics in the UK, for GEC Australia and was the founding Managing Director of the Automation and Engineering Applications Centre Ltd which designed and built industrial automation systems. He recently retired after 14 years as Executive Dean of Engineering and Information Sciences at the University of Wollongong but remains an active researcher in robotics and machine intelligence.  
\end{IEEEbiography}

\begin{IEEEbiography}[{\includegraphics[width=1in,height=1.25in,clip,keepaspectratio]{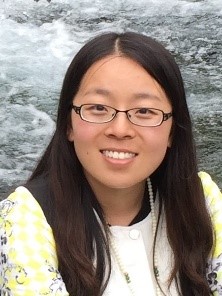}}]{Yanbo Gao} 
is currently with the School of Software, Shandong University (SDU), Jinan, China, as an Associate Professor. She was with the School of Information and Communication Engineering, University of Electronic Science and Technology of China (UESTC), Chengdu, China, as a Post-doctor from 2018-2020. She received her Ph.D. degree from UESTC in 2018. Her research interests include video coding, 3D video processing and light field image coding. She was a co-recipient of the best student paper awards at the IEEE BMSB 2018.
\end{IEEEbiography}

\vfill
\begin{figure*}[tbp]
	\centering
	\includegraphics[width=1\hsize]{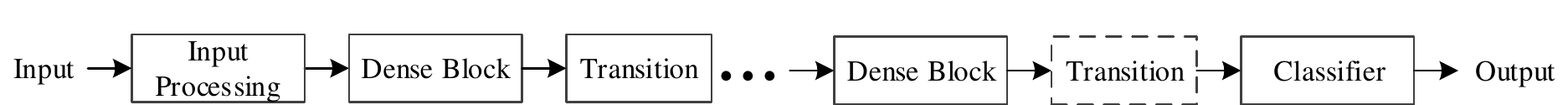}
	\caption{Framework of the densely connected IndRNN used in the experiments.} \label{fig_framework_denseIndRNN}
\end{figure*}

\newpage

\appendices
\section{Relationship Between IndRNN and RNN}
\label{appenx_a}
The relationship between IndRNN and the traditional RNN is illustrated in the following, where we show that under certain circumstances the traditional RNN is only a special case of a two-layers IndRNN. First, from the perspective of number of neurons, as shown in \ref{sec_comp_neuron}, for a $N$-neuron RNN network with input of dimension $M$, the number of parameters using traditional RNN is $M\times N+N\times N$, while the number of parameters using one-layer IndRNN is $M\times N+N$. For a two-layers IndRNN where both layers consist of $N$ neurons, the number of parameters is $M\times N+N\times N+2\times N$, which is of a similar order to the traditional RNN. Therefore, in the following we compare a two-layer IndRNN with a single layer RNN. For simplicity, the bias term is ignored for both IndRNN and traditional RNN. 

Assume a simple $N$-neuron two-layer network where the recurrent weights for the second layer are zero which means the second layer is just a fully connected layer shared over time. The Hadamard product ($\mathbf{u}\odot\mathbf{h}_{t-1}$) can be represented in the form of matrix product by $diag(u_{1}, u_{2}, \hdots, u_{N})\mathbf{h}_{t-1}$. In the following, $diag(u_{1}, u_{2}, \hdots, u_{N})$ is shortened as $diag(u_{i})$. Assume that the activation function is a linear function $\sigma (x)=x$. The first and second layers of a two-layer IndRNN can be represented by (\ref{firstlayerKIRNN}) and (\ref{secondlayerKIRNN}), respectively. 
\begin{align}
\label{firstlayerKIRNN}
\mathbf{h}_{f,t}&=\mathbf{W}_f\mathbf{x}_{f,t}+diag(u_{fi})\mathbf{h}_{f,t-1}
\\
\label{secondlayerKIRNN}
\mathbf{h}_{s,t}&=\mathbf{W}_s\mathbf{h}_{f,t}
\end{align}
Assuming $\mathbf{W}_s$ is invertible, then 
\begin{equation}
\mathbf{W}_s^{-1}\mathbf{h}_{s,t}=\mathbf{W}_f\mathbf{x}_{f,t}+diag(u_{fi})\mathbf{W}_s^{-1}\mathbf{h}_{s,t-1}
\end{equation}
Thus
\begin{equation}
\mathbf{h}_{s,t}=\mathbf{W}_s\mathbf{W}_f\mathbf{x}_{f,t}+\mathbf{W}_sdiag(u_{fi})\mathbf{W}_s^{-1}\mathbf{h}_{s,t-1}
\label{kirnnreprnn}
\end{equation}

 By assigning $\mathbf{U}=\mathbf{W}_sdiag(u_{fi})\mathbf{W}_s^{-1}$ and $\mathbf{W}=\mathbf{W}_s\mathbf{W}_f$, it becomes
\begin{equation}
\mathbf{h}_t=\mathbf{Wx}_t+\mathbf{Uh}_{t-1}
\label{nobrnn}
\end{equation}
which is a traditional RNN. Note that this only imposes the constraint that the recurrent weight ($\mathbf{U}$) is diagonalizable. Therefore, the simple two-layer IndRNN network can represent a traditional RNN network with a diagonalizable recurrent weight ($\mathbf{U}$). In other words, under linear activation, a traditional RNN with a diagonalizable recurrent weight ($\mathbf{U}$) is a special case of a two-layer IndRNN. 

It is known that a non-diagonalizable matrix can be made diagonalizable with a perturbation matrix composed of small entries. A stable RNN network needs to be robust to small perturbations (in order to deal with precision errors for example). It is possible to find an RNN network with a diagonalizable recurrent weight matrix to approximate a stable RNN network with a non-diagonalizable recurrent weight matrix. Therefore, a traditional RNN with a linear activation is a special case of a two-layer IndRNN. For a traditional RNN with a nonlinear activation function, its relationship with the proposed IndRNN is yet to be established theoretically. However, we have shown empirically that the proposed IndRNN can achieve better performance than a traditional RNN with a nonlinear activation function in the experiments. Therefore, the cross-channel information can be well explored with a multiple-layer IndRNN although IndRNN neurons are independent of each other in each layer. In all, multiple-layer IndRNNs can be used in place of the traditional RNNs.

\begin{figure}[tbp]
	\centering
	\includegraphics[width=1\hsize]{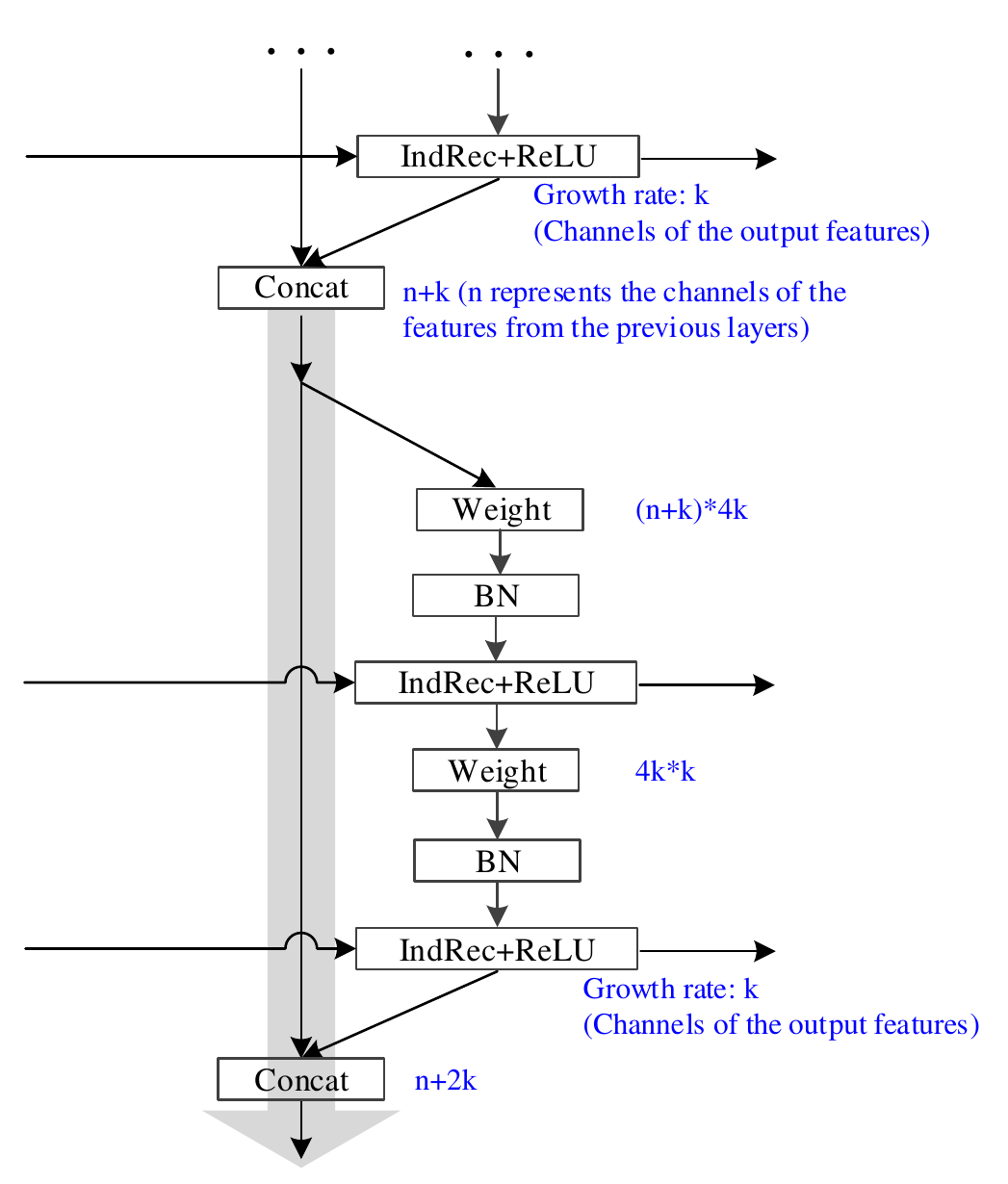}
	\caption{Detailed illustration of the dense IndRNN block.} \label{fig_denseblock}

\end{figure}

\section{Description of the Densely Connected IndRNN Setting}
\label{appenx_b}
The framework of the densely connected IndRNN process is shown in Fig. \ref{fig_framework_denseIndRNN}, which generally follows the conventional denseCNN \cite{huang2017densely}. First, the input is processed into features usually with the embedding layer (in the language modelling task) or one IndRNN layer (in the other recognition tasks). Then the initial features are processed with several dense IndRNN blocks and transition layers. As stated in Subsection \ref{subsec_trainsetup}, three dense IndRNN blocks are used with 8, 6 and 4 dense IndRNN layers, respectively. While in the conventional denseCNN \cite{huang2017densely} the last transition layer is not needed after the final dense block, we find that it usually improves the performance in our experiments. Therefore, the last transition layer is also added in our dense-IndRNN. Finally, a classifier is added in the end for recognition. 

Fig. \ref{fig_denseblock} further shows the details of the dense IndRNN block. Each dense IndRNN layer produces $k$ channels of output features, termed as growth rate. After one dense IndRNN layer, the features are increased to $n+k$ channels, where $n$ represents the channels of the features from the previous layers. For the next IndRNN layer, the first weight processes the $n+k$ dimensional to a $4k$-channel feature, then further processed by BN and IndRec+ReLU. The second weight processes the $4k$-channel feature to $k$-channel and further processed for output. Concatenated with the input $n+k$-channel feature, the final output feature of the dense IndRNN layer becomes $n+2k$ channels. For the transition layers, it processes the feature ($N$ channels) from the dense IndRNN block and reduces it to half ($N/2$ channels). The key component of dense-IndRNN different from the residual-IndRNN is the concatenation operation of the dense layers and the shortcut, which allows the features to be reused in the following layers.

\end{document}